\documentclass[journal]{IEEEtran}
\ifCLASSINFOpdf
\else
\fi
\usepackage{ulem}
\usepackage{float}
\usepackage{algorithm}
\usepackage{algpseudocode}
\usepackage[marginal]{footmisc}
\usepackage[numbers,square,comma,sort&compress]{natbib}

\usepackage{graphics} 
\usepackage{epsfig} 
\usepackage{mathptmx} 
\usepackage{times} 
\usepackage[table]{xcolor}  
\usepackage{colortbl}
\usepackage{indentfirst}
\usepackage{amsmath}
\usepackage{bbding}
\usepackage{amssymb}
\usepackage{pifont}
\usepackage{amssymb}  
\usepackage{newtxtext}
\usepackage{newtxmath}
\usepackage{booktabs}
\usepackage{url}
\usepackage{multirow}
\usepackage{bbding}
\usepackage{hyperref}
\usepackage{xcolor}
\usepackage{blindtext}
\DeclareMathAlphabet{\mathcal}{OMS}{cmsy}{m}{n}

\usepackage{etoolbox}
\makeatletter
\patchcmd{\@makecaption}
  {\scshape}
  {}
  {}
  {}
\makeatletter
\patchcmd{\@makecaption}
  {\\}
  {.\ }
  {}
  {}
\makeatother

\newcommand{\xmark}{\ding{55}}

\begin{document}
%
\title{Talk2Sensors: 3D Visual Grounding in Autonomous Driving via Sensor-Adaptive Physical Cue Matching}
%
%
%

\author{Runwei Guan$^\dagger$,
        Di Tian$^\dagger$,
        Ningwei Ouyang,
        Ruixiao Zhang,
        Shaofeng Liang,
        Haocheng Zhao,
        Lianqing Zheng,
        Xiaokai Bai,
        Guotao Wang,
        Daizong Liu$^*$,
        Henghui Ding,
        and~Hui~Xiong$^*$,~\IEEEmembership{Fellow,~IEEE}
\thanks{$^\dagger$Runwei Guan and Di Tian are co-first authors.}
\thanks{$^*$Corresponding authors: daizongliu@whu.edu.cn, xionghui@hkust-gz.edu.cn.}
\thanks{Runwei Guan, Shaofeng Liang and Hui Xiong are with Thrust of Artificial Intelligence, The Hong Kong University of Science and Technology (Guangzhou). Runwei Guan is also with MMLab, CUHK. Hui Xiong is also with Department of CSE, Hong Kong SAR. Email: \{runwayrwguan, xionghui\}@hkust-gz.edu.cn, shaofengliang1999@163.com}
\thanks{Di Tian is with School of Transportation Science and Engineering, Harbin Institute of Technology. Email: 18753017323@163.com.}
\thanks{Ningwei Ouyang is with School of Advanced Technology, Xi'an Jiaotong-Liverpool University, Suzhou, China. Email: psnouyan@liverpool.ac.uk.}
\thanks{Ruixiao Zhang is with School of Electronics and Computer Science, University of Southampton. Email: rz3e25@soton.ac.uk.}
\thanks{Haocheng Zhao is with School of Intelligent Manufacturing and Smart Transportation, Suzhou City University, Suzhou, China. Email: Haocheng.Zhao@szcu.edu.cn.}
\thanks{Guotao Wang is with Qingdao University of Science and Technology. Email: qduwgt@163.com.}
\thanks{Daizong Liu is with Institute for Math \& AI, Wuhan University. Email: daizongliu@whu.edu.cn.}
\thanks{Henghui Ding is with Institute of Big Data, Fudan University. Email: henghui.ding@gmail.com.}
}

%
%

\markboth{}%
{Shell \MakeLowercase{\textit{et al.}}: Bare Demo of IEEEtran.cls for IEEE Journals}
%



\maketitle

\begin{abstract}
As a key capability for embodied intelligence, 3D visual grounding (3DVG) has been predominantly studied in indoor scenes with RGB-D or point-cloud inputs, while existing outdoor extensions largely rely on monocular images alone. Both settings fall short of real-world outdoor perception, where heterogeneous sensors capture complementary yet distinct physical properties, such as visual texture, 3D geometry, and object kinematics, that are indispensable for flexible and robust query-adaptive grounding but remain under-exploited. To bridge this gap, we introduce Talk2Sensors, the first multi-sensor 3D visual grounding dataset built upon camera, LiDAR, and 4D radar. It contains 8,682 language instructions and 20,558 referred objects, with diverse prompts explicitly aligned with sensor-specific physical cues.
Furthermore, we propose TSFormer, a unified Transformer-based framework for language-guided 3D visual grounding in autonomous driving. TSFormer adopts a coarse-to-fine property-aware fusion strategy: the Language-Routed Property Sampler first performs coarse text-conditioned feature retrieval by modulating sensor sampling weights with query-level linguistic cues, while the subsequent Sparse-Preserving Modality Arbiter module conducts fine-grained modality arbitration and text-guided refinement to determine the precise referred spatial location. This design enables dynamic routing of appearance, geometry, and motion cues according to the semantic requirements of each prompt, preventing dense modalities from overwhelming sparse but critical sensor signals. Extensive experiments demonstrate that TSFormer achieves state-of-the-art performance across multiple benchmarks: it improves over the strongest baseline by 8.05 mAP on Talk2Sensors, and transfers to the monocular Mono3DRefer benchmark with 53.05\% Acc@0.5, establishing a strong foundation for multi-sensor interactive scene understanding. The project is available at \href{https://github.com/GuanRunwei/Talk2Sensors}{here}.
\end{abstract}


%
\IEEEpeerreviewmaketitle

\section{Introduction}
%
%
%
%
\IEEEPARstart{I}{n} the rapidly evolving domains of embodied artificial intelligence (AI) and interactive scene understanding, 3D visual grounding, also referred to as 3D referring expression comprehension, has emerged as a foundational capability \cite{liu2025survey}. Unlike traditional object detection that rigidly confines perception to predefined categories, 3D visual grounding bridges the critical semantic gap between human instructions and the physical world, enabling machines to interpret and execute open-ended, fine-grained user intents. Typically, this task utilizes natural language text as semantic guidance \cite{ren2026mtrag}, directing perception systems to reliably retrieve and localize one or multiple referred objects within complex 3D scene contexts \cite{guo2025visual}. Consequently, this capability serves as an indispensable prerequisite for advanced downstream autonomous operations, facilitating safe path planning, navigation, and robust human-computer interaction for diverse intelligent platforms such as autonomous surface vehicles and drones \cite{liu2025embodied}.

\begin{figure}
    \includegraphics[width=0.99\linewidth]{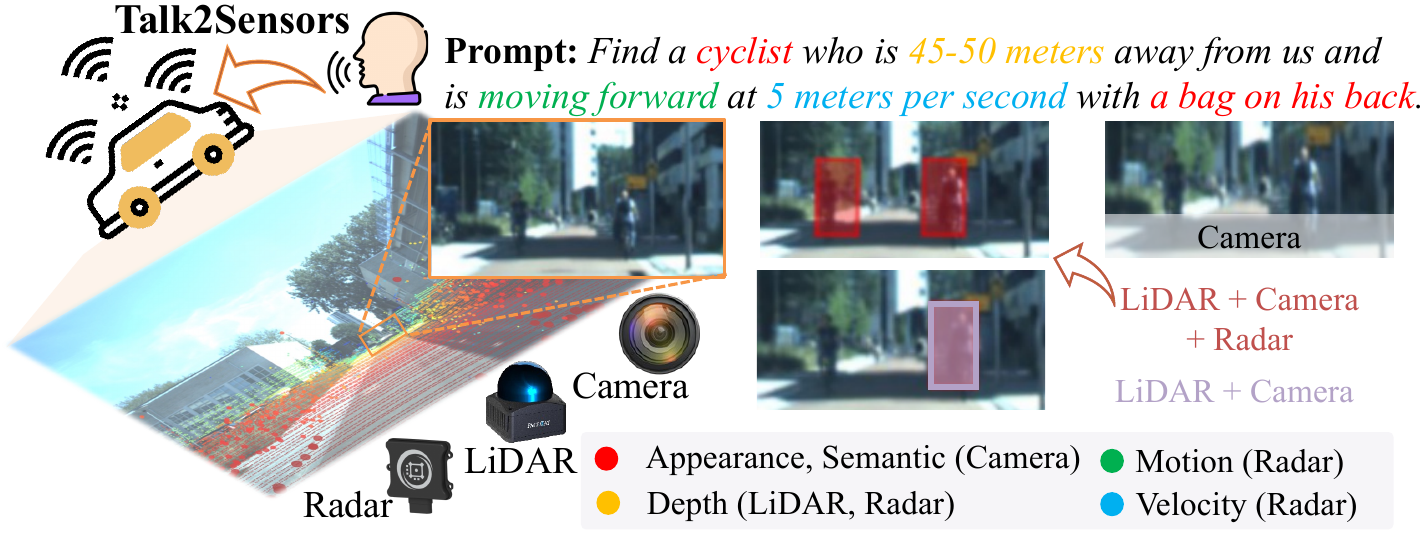}
    \vspace{-3mm}
    \caption{The overview of Talk2Sensors pipeline, regarding textual prompt-guided multi-sensor 3D visual grounding.}
    \label{fig:overview}
\end{figure}

\begin{table*}
\centering
\caption{Comparison of Talk2Sensors with existing datasets. Our dataset is the first to support language-guided 3D visual grounding with tri-sensor (camera, LiDAR, radar) inputs and explicit physical property alignment.}
\vspace{-3mm}
\resizebox{\textwidth}{!}{
\begin{tabular}{lccccccccc}
\toprule
\textbf{Dataset} & \textbf{Scene} & \textbf{Venues} & \textbf{Cam.} & \textbf{LiDAR} & \textbf{4D Radar} &  \textbf{Property-Aware} & \textbf{\#Samples} & \textbf{\#Objects} \\
\midrule
ScanRefer \cite{chen2020scanrefer}  & Indoor  & ECCV$_{2020}$ &  \checkmark & \checkmark & \xmark &  \xmark & 51K  & 11K \\
SUNRefer \cite{liu2021refer} & Indoor  & CVPR$_{2021}$  & \checkmark & \checkmark & \xmark &  \xmark & 20K  & 8K  \\
ReferIt3D \cite{achlioptas2020referit3d}        & Indoor  & ECCV$_{2020}$ & \checkmark & \checkmark & \xmark &  \xmark & 13K  & 6K  \\
Talk2Car \cite{deruyttere2019talk2car}         & Outdoor & EMNLP$_{2019}$  & \checkmark & \xmark & \xmark &  \xmark & 10K  & -   \\
Talk2Radar \cite{guan2025talk2radar}      & Outdoor & ICRA$_{2025}$  & \xmark & \checkmark & \checkmark &  \xmark & 8.7K   & -   \\
Mono3DRefer \cite{zhan2024mono3dvg} & Outdoor & AAAI$_{2024}$ & \checkmark & \xmark & \xmark & \xmark & 2.0k & 8.2k  \\
\midrule
\textbf{Talk2Sensors (Ours)} 
                 & Outdoor  & - & \checkmark & \checkmark & \checkmark &  \checkmark & 8.7K & 20.6K \\
\bottomrule
\end{tabular}
}
\label{tab:dataset_comparison}
\end{table*}

Driven by the rapid advancements in deep learning and vision-language models (VLMs) \cite{chung2026enhanced}, 3D visual grounding has experienced significant progress. Current research predominantly focuses on both indoor and outdoor scenarios, delving into increasingly complex referring objects. These include reasoning based on fine-grained appearances, describing the relative spatial relationships among multiple entities, and resolving references with quantitative attributes \cite{li2025seeground,guan2025talk2pc}. However, existing methods overlook the fact that different sensors inherently capture distinct physical properties of objects \cite{zhan2024mono3dvg,chen2020scanrefer,liu2021refer}. These physical characteristics serve as a critical bottleneck that restricts the upper bound of textual referring capabilities \cite{hou2026mmdrive}. For instance, if a 3D visual grounding system relies solely on an RGB camera as input, querying for ``a pedestrian approximately 30 meters away moving at roughly 2 m/s" is fundamentally unfeasible, as the camera lacks the capability to perceive depth and kinematic information. Consequently, it is imperative to investigate 3D visual grounding frameworks that utilize multi-sensor inputs to represent the scene. Such an approach diversifies the referring attributes within the text and enables semantic-driven, dynamic querying of heterogeneous physical features captured by diverse sensors, as Fig. \ref{fig:overview} shows. Ultimately, this maximizes the capability of 3D visual grounding systems to recognize and localize objects under varying physical constraints.

Motivated by the above, \textbf{(1)} We introduce Talk2Sensors, a novel dataset dedicated to multi-sensor 3D visual grounding. Built upon three sensors widely deployed in autonomous driving: RGB cameras, LiDAR, and 4D millimeter-wave (mmWave) radar, Talk2Sensors aims to establish a flexible and diverse benchmark by leveraging the distinct physical properties inherently captured by each modality. Specifically, the referring expressions in our dataset incorporate partial or comprehensive descriptions spanning three representative feature categories: visual characteristics (e.g., color, texture, and pattern), LiDAR-derived attributes (e.g., 3D contour, shape, and geometric structure), and mmWave radar properties (e.g., motion tendency, velocity, penetrability in adverse weather, and radar cross-section). By seamlessly aligning linguistic semantics with these heterogeneous physical traits, this benchmark shifts the research paradigm from purely context-driven referring to comprehensive, physics-aware scene understanding. Exactly, as Table \ref{tab:dataset_comparison} shows, Talk2Sensors includes 8,682 samples with referred expressions, containing 20,558 referred objects with three primary categories. 

\textbf{(2)} We propose TSFormer, a unified language-guided framework tailored to this tri-sensor, property-aware setting. Directly adapting existing methods to Talk2Sensors is non-trivial, as they suffer from three key limitations. \textbf{First}, most multi-sensor detectors fuse modalities through rigid, calibration-dependent geometric projection into a shared BEV or voxel space~\cite{liu2023bevfusion, chen2023futr3d}; such hand-crafted alignment is brittle across the drastically different densities of camera, LiDAR, and radar, and does not readily generalize to arbitrary sensor subsets. \textbf{Second}, prevailing fusion schemes (e.g., concatenation or summation~\cite{arevalo2017gated, hu2018squeeze}) are agnostic to the query, so information-dense camera features tend to overwhelm the sparse yet decisive radar signals that encode motion and velocity. \textbf{Third}, existing 3DVG models~\cite{guan2024talk2radar, guan2025talk2pc, zhan2024mono3dvg} ground on contextual appearance and fuse modalities in a language-agnostic manner, lacking any mechanism to route a prompt toward the sensor that actually carries the queried physical attribute (e.g., geometry from LiDAR, kinematics from radar). To resolve these issues, TSFormer introduces a Language-Routed Property Sampler that projects heterogeneous sensors into a unified query space~\cite{wang2022detr3d} and performs coarse text-conditioned retrieval, thereby circumventing heuristic geometric alignment, together with a Sparse-Preserving Modality Arbiter module that dynamically arbitrates modalities under linguistic guidance, preventing dense signals from overwhelming sparse but critical cues and achieving robust, query-adaptive cross-modal grounding.

In summary, the main contributions of this paper are summarized as follows:
\begin{itemize}
    \item \textbf{A Novel Multi-Sensor Benchmark:} We introduce \textbf{Talk2Sensors}, the first 3D visual grounding dataset built upon camera, LiDAR, and 4D mmWave radar. It contains 8,682 prompts and 20,558 referred objects, explicitly aligning language descriptions with heterogeneous physical cues.

    \item \textbf{A Unified Tri-Sensor Grounding Framework:} We propose \textbf{TSFormer}, a language-guided framework for multi-sensor 3D visual grounding. It unifies camera, LiDAR, and radar features in a query-based architecture, enabling flexible sensor feature retrieval without hand-crafted fusion rules.

    \item \textbf{A Coarse-to-Fine Property-Aware Fusion Mechanism:} We design a Language-Routed Property Sampler followed by a Sparse-Preserving Modality Arbiter. The sampler coarsely retrieves text-relevant sensor cues, while the fusion module adaptively emphasizes appearance, geometry, or motion information according to the prompt, achieving robust cross-modal grounding.
\end{itemize}

The remainder of this paper is organized as follows. Section~\ref{sec:related} reviews related work on multi-sensor fusion and 3D visual grounding. Section~\ref{sec:dataset} presents the proposed Talk2Sensors dataset, and Section~\ref{sec:method} details the TSFormer framework. Section~\ref{sec:experiments} reports experimental results and analyses, and Section~\ref{sec:conclusion} concludes the paper.

\begin{figure*}
    \includegraphics[width=0.99\linewidth]{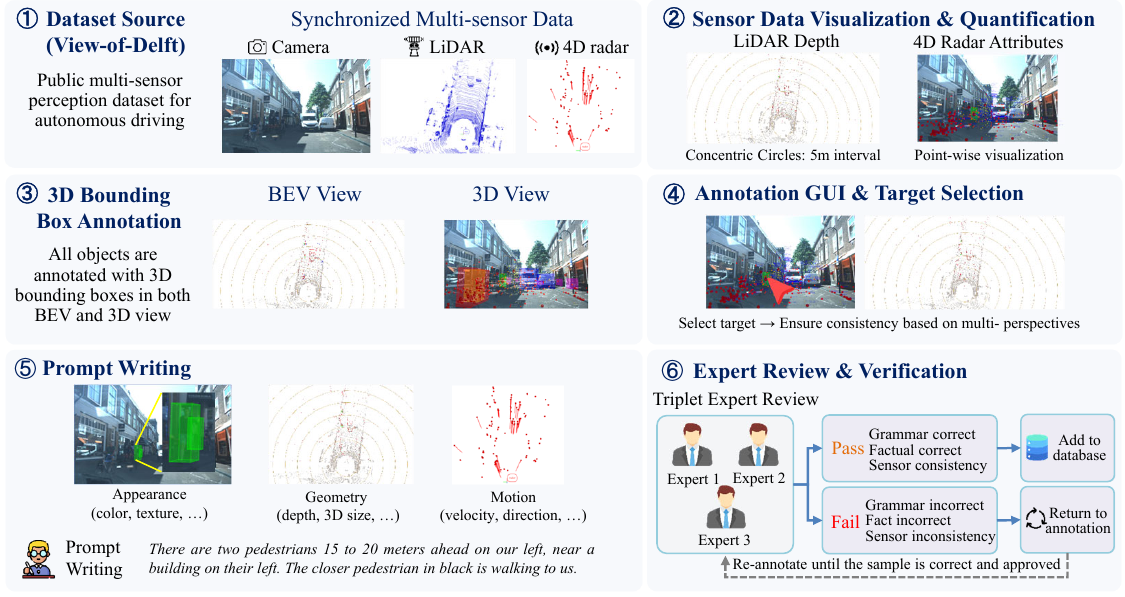}
    \vspace{-3mm}
    \caption{The annotation process of our proposed Talk2Sensors dataset. Following a pipeline of quantification-annotation-verification, implicit sensor attributes are first visualized as quantifiable cues, upon which annotators label 3D boxes and write property-grounded prompts (appearance, geometry, motion); each sample is then verified by three experts for factual and multi-sensor consistency.}
    \label{fig:annotation}
\end{figure*}

\section{Related Works}\label{sec:related}
\subsection{Fusion of Camera, LiDAR and mmWave Radar for Autonomous Driving}
Multi-sensor fusion is an indispensable component of robust autonomous driving \cite{wang2026rolic,zheng2026omnihd}, significantly reducing perception uncertainty by leveraging the complementary physical properties of cameras, LiDAR, and mmWave radar. While cameras provide dense semantic information such as color and texture, they lack reliable depth estimation. Conversely, LiDAR captures high-resolution 3D geometric structures but suffers severe degradation in adverse weather, a vulnerability effectively mitigated by mmWave radar, which yields robust kinematic measurements (e.g., velocity) regardless of environmental visibility. To harness these heterogeneous modalities, recent studies have proposed various advanced fusion paradigms. FUTR3D \cite{chen2023futr3d} introduces a highly flexible, end-to-end framework utilizing a query-based modality-agnostic feature sampler to dynamically extract features across arbitrary sensor configurations, circumventing complex heuristic late-fusion designs. Similarly, ASF \cite{paek2026availability} proposes an availability-aware fusion mechanism that aligns heterogeneous data into a unified canonical space, applying cross-attention to adaptively adjust sensor weights in response to hardware degradation. Furthermore, SAMFusion \cite{palladin2024samfusion} specifically targets adverse weather scenarios by employing a depth-based adaptive blending module and a distance-aware multimodal proposal network, strategically balancing LiDAR and radar based on their optimal operating ranges.

Despite progress in spatial alignment and physical robustness \cite{fu2025boosting}, existing methods primarily rely on heuristic geometric cues, largely overlooking the critical role of high-level semantic intent in guiding multi-sensor perception \cite{liu2024talk,guan2025talk2radar,baek2024lidarefer}. To address this limitation in 3D visual grounding, we propose a novel fusion paradigm explicitly driven by natural language. Our approach utilizes linguistic descriptions, such as color, geometry, or velocity, to dynamically route attention toward the most relevant sensor modality. Furthermore, we introduce a soft-gated sparse cross-attention mechanism that effectively prevents information-dense modalities (e.g., RGB images) from overwhelming inherently sparse, yet critical, radar signals. Ultimately, this semantic-driven routing maximizes the utility of heterogeneous physical properties, achieving a highly robust and query-adaptive cross-modal alignment.

\begin{table*}
\caption{Statistics of referent object number and point clouds in Talk2Sensors}
\vspace{-3mm}
\label{tab:t2s_stats}
\centering
\footnotesize
\setlength{\tabcolsep}{5.5pt}
\begin{tabular}{@{}c|*{12}{c}@{}}
  \toprule
  \multirow{2}{*}{\textbf{Objects}} & \multirow{2}{*}{\textbf{Pedestrian}} & \multirow{2}{*}{\textbf{Cyclist}} & \multirow{2}{*}{\textbf{Car}}
    & \multirow{2}{*}{\textbf{Motor}} & \multirow{2}{*}{\textbf{Truck}} & \multirow{2}{*}{\textbf{Bicycle}} & \multirow{2}{*}{\textbf{Rider}}
    & \textbf{\shortstack{Moped}} & \textbf{\shortstack{Bicycle}}
    & \textbf{\shortstack{Human}} & \textbf{\shortstack{Rider}}
    & \textbf{\shortstack{Vehicle}}\\
    & & & & & & & & \textbf{Scooter} & \textbf{Rack} & \textbf{Depiction} & \textbf{other} & \textbf{other} \\
  \midrule
  Sensor PC
    & \shortstack{3487\\(16.96\%)} & \shortstack{2157\\(10.49\%)} & \shortstack{9336\\(45.41\%)}
    & \shortstack{73\\(0.36\%)}   & \shortstack{27\\(0.13\%)}   & \shortstack{2442\\(11.88\%)}
    & \shortstack{1236\\(6.01\%)} & \shortstack{470\\(2.29\%)}  & \shortstack{1165\\(5.67\%)}
    & \shortstack{48\\(0.23\%)}   & \shortstack{11\\(0.05\%)}   & \shortstack{3\\(0.01\%)}\\
  \midrule
  Radar 1 & 1.3   & 9.3   & 7.4   & 0.2  & 10.2   & 0.7   & 3.7   & 5.9   & 1.7   & 1.4   & 0    & 0     \\
  Radar 3 & 4.2   & 11.3  & 11.4  & 9.2  & 56.7   & 3.6   & 4.9   & 4.8   & 9.0   & 0.7   & 8.6  & 9.7   \\
  Radar 5 & 5.9   & 14.8  & 18.3  & 11.9 & 90.9   & 6.0   & 6.1   & 7.1   & 14.8  & 1.3   & 13.0 & 17.3  \\
  LiDAR   & 107.8 & 220.9 & 368.3 & 333.8& 2330.0 & 106.7 & 106.2 & 129.6 & 204.2 & 204.7 & 256.2& 165.3 \\
  \bottomrule
\end{tabular}
\end{table*}

\subsection{3D Visual Grounding in Autonomous Driving Scenarios}
Visual grounding aims to establish fine-grained correspondences between natural language descriptions and specific visual elements in an image or 3D scene \cite{liu2025survey}. In autonomous driving scenarios, 3D visual grounding extends this capability to LiDAR point clouds or camera data, enabling vehicles to interpret high-level linguistic commands (e.g., ``the red car parked under the tree on the left”) and localize target objects in complex, dynamic outdoor environments.

Autonomous driving introduces unique challenges: large-scale outdoor LiDAR scenes dominated by background points, dynamic objects, multi-sensor fusion (LiDAR, radar, RGB camera), and the necessity for real-time, context-aware reasoning \cite{wang2023multi}.
Early efforts extended 2D grounding to driving contexts. The Talk2Car dataset \cite{deruyttere2019talk2car} and its 3D extension Talk2Car-3D \cite{cheng2023language} have become standard benchmarks. Context-aware models such as CAVG \cite{liao2024gpt} integrate GPT-4 for multimodal reasoning, combining text, emotion, image, and cross-modal encoders to handle ambiguous driving commands.
For 3D visual grounding in outdoor scenes, recent works have made significant progress. LidaRefer \cite{baek2024lidarefer} is a context-aware transformer-based framework designed for outdoor LiDAR data while Talk2LiDAR \cite{liu2024talk} builds a benchmarks upon LiDAR for 3D visual grounding. 
Building on multi-view data, NuGrounding \cite{li2025nugrounding} introduces the first large-scale benchmark for multi-view visual grounding in autonomous driving.
Moreover, Talk2Radar \cite{guan2025talk2radar} introduces the first mmWave radar-centric 3D visual grounding dataset, leveraging radar-captured physical attributes to investigate the modality's unique advantages. Based on Talk2Radar dataset, TPCNet fuses LiDAR and radar by proposed Dynamic Gated Graph Fusion (DGGF) to obtain more precise results for 3D visual grounding \cite{guan2025talk2pc}. Besides, zero-shot approaches such as VLM-Grounder \cite{xu2024vlm} further leverage 2D vision-language models for 3D grounding via multi-view stitching and ensemble projection, reducing reliance on expensive 3D annotations.

Despite these remarkable advancements, existing benchmarks and models primarily focus on single or dual modalities \cite{guan2024talk2radar,guan2025talk2pc}, failing to fully exploit the heterogeneous physical properties simultaneously captured by cameras, LiDAR, and 4D mmWave radar (e.g., visual texture, precise geometry, and kinematics). To bridge this critical gap, we pioneer Talk2Sensors, the first comprehensive tri-sensor 3D visual grounding dataset. By explicitly aligning natural language prompts with distinct sensor-specific physical attributes, Talk2Sensors shifts the research paradigm toward property-aware scene understanding.

\begin{figure*}
    \includegraphics[width=0.99\linewidth]{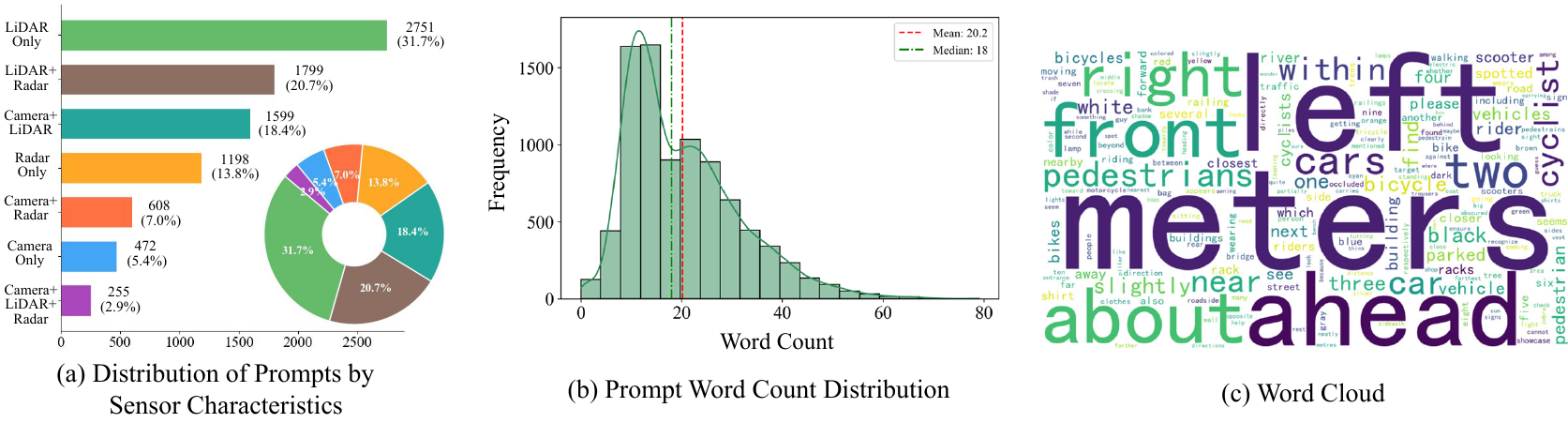}
    \vspace{-3mm}
    \caption{Statistical overview of the Talk2Sensors dataset. (a) Proportion of queries relying on different sensor configurations. (b) Distribution of prompt lengths, demonstrating the complexity of the instructions. (c) Word cloud highlighting the prevalent usage of geometric and quantitative physical attributes.}
    \label{fig:stat_1}
\end{figure*}

\section{The Proposed Dataset}\label{sec:dataset}
To bridge the critical gap between abstract linguistic concepts and the heterogeneous physical world, we introduce Talk2Sensors, a pioneering multi-sensor 3D visual grounding dataset that redefines how autonomous systems perceive and reason about their surroundings. Unlike existing benchmarks that predominantly rely on context-driven reasoning within single or dual modalities, Talk2Sensors is the first to explicitly align natural language instructions with the complementary physical properties inherently captured by cameras, LiDAR, and 4D mmWave radar, namely visual texture, 3D geometry, and absolute kinematics. Comprising 8,682 diverse instructions and 20,558 referred objects, this dataset uniquely features a property-aware annotation paradigm where linguistic queries are dynamically grounded in the specific physical traits provided by tri-sensor inputs. By establishing this comprehensive physical-semantic mapping, Talk2Sensors shifts the research focus from purely contextual inference toward robust, physics-aware scene understanding in complex autonomous driving scenarios.

\subsection{The Annotation Process}
To construct a high-quality, physics-aligned 3D visual grounding dataset, we designed a rigorous closed-loop pipeline of quantification-annotation-verification (Fig. \ref{fig:annotation}). This process ensures the absolute objectivity of the linguistic prompts and their precise alignment with heterogeneous sensor measurements. The pipeline consists of four primary stages:

\subsubsection{Data Sourcing and Multi-modal Alignment}
The raw data is sourced from the View-of-Delft (VoD) dataset, a specialized multi-sensor perception benchmark for autonomous driving. VoD provides strictly synchronized and calibrated data from RGB cameras, LiDAR, and 4D mmWave radar. This tri-sensor synchronization provides a reliable foundation for extracting and fusing complex physical features.

\subsubsection{Quantitative Feature Visualization}
To eliminate subjective spatial bias during human annotation, we explicitly visualize implicit physical attributes as quantifiable visual cues. For the 4D mmWave radar data, we project point-wise attributes, including radial compensated velocity, depth, azimuth, and Radar Cross-Section (RCS), directly into the visual workspace. For LiDAR, depth information is visualized as a point-wise gradient. Furthermore, we render concentric circles at 5-meter intervals centered on the ego-vehicle in both the Bird’s-Eye View (BEV) and 3D views. This rigorous quantitative reference frame allows annotators to distinguish object depth and azimuth with sub-meter precision, stripping away the errors inherent in human visual estimation.

\subsubsection{Objective Prompt Generation}
All potential objects in the scene are pre-labeled with 3D bounding boxes. Utilizing our custom-developed Annotation GUI, annotators can anchor a specific object by clicking, which highlights the intended object and filters out distracting background elements. Each referring prompt is then manually written by trained human annotators for the anchored object, rather than generated from templates or large language models, ensuring naturalistic and unambiguous descriptions. During prompt writing, we impose a strict physical-only constraint: descriptions must be strictly derived from observable physical facts, such as visual appearance (camera), 3D geometric structure (LiDAR), and absolute kinematics (radar). Any subjective inferences, ambiguous qualifiers, or emotional descriptions are prohibited to ensure that linguistic semantics are grounded in the underlying sensor measurements.

\subsubsection{Triplet Expert Review and Verification}
To guarantee the ultimate integrity of the dataset, every text-object pair undergoes a multi-tier quality assurance protocol. We introduce three independent experts in the field of autonomous driving to perform a cross-validation of the samples. The review process evaluates each sample across three critical dimensions: grammatical correctness, factual accuracy, and multi-sensor physical consistency. A sample is admitted into the final database only upon unanimous approval by all three experts. Samples that exhibit semantic ambiguity or physical inconsistencies are rejected and returned for re-annotation until the rigorous criteria are met. This closed-loop mechanism fundamentally eliminates semantic noise and ensures a robust physical-semantic mapping.

\begin{figure}
    \includegraphics[width=0.99\linewidth]{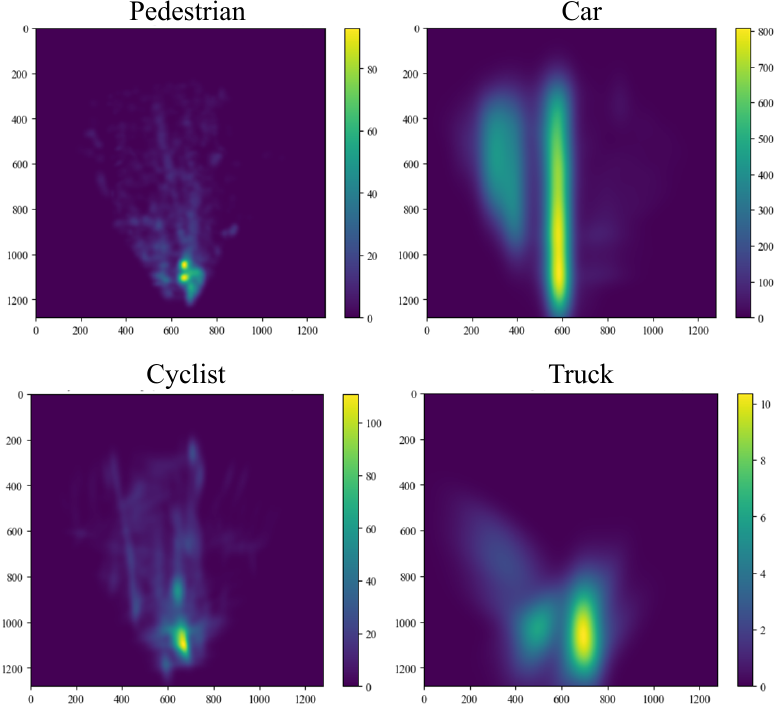}
    \vspace{-3mm}
    \caption{Spatial distribution (BEV) of referred objects in Talk2Sensors. The BEV heatmaps depict the frequency and spatial spread of Pedestrians, Cars, Cyclists, and Trucks relative to the ego-vehicle, showcasing the spatial diversity and complexity of the dataset.}
    \label{fig:dataset_stats_2}
\end{figure}

\subsection{Dataset Statistics}
To demonstrate the diversity, complexity, and unique multi-modal nature of Talk2Sensors, we comprehensively analyze its statistical properties from three primary perspectives: prompt complexity, sensor-adaptive distribution, and spatial object diversity, which are presented in Fig. \ref{fig:stat_1}.

\begin{figure*}
    \includegraphics[width=0.99\linewidth]{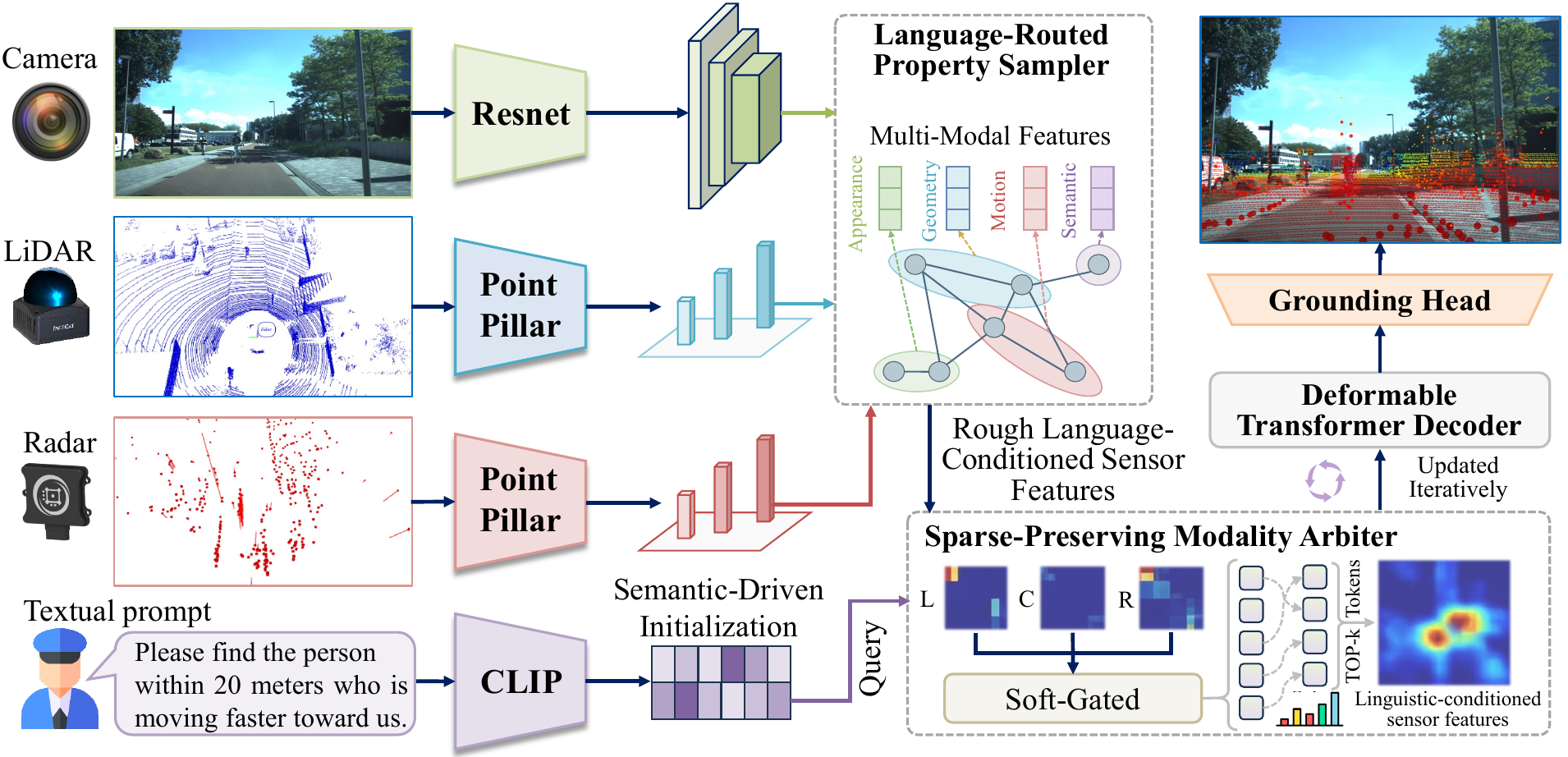}
    \vspace{-3mm}
    \caption{The pipeline of TSFormer for multi-sensor 3D visual grounding. The framework comprises a Language-Routed Property Sampler for semantic-driven initial anchoring and feature retrieval, followed by a Sparse-Preserving Modality Arbiter to dynamically filter and fuse heterogeneous physical cues (Appearance, Geometry, Motion) under strict linguistic guidance, achieving precise 3D localization.}
    \label{fig:pipeline}
\end{figure*}

\subsubsection{Prompt Complexity and Semantic Richness}
The dataset comprises 8,682 linguistically diverse instructions. As illustrated in the prompt length distribution, the word count exhibits a right-skewed normal distribution with a mean of 20.2 words and a median of 18 words. This average length significantly surpasses traditional visual grounding benchmarks, indicating a higher density of physical constraints and attribute descriptions. Furthermore, the word cloud reveals a strong linguistic emphasis on quantitative spatial measurements (e.g., ``meters") and precise directional relations (e.g., ``left", ``right", ``ahead"), which strictly require models to possess robust 3D geometric reasoning capabilities.

\subsubsection{Sensor-Adaptive Prompt Distribution}
Unlike existing datasets that predominantly rely on context-driven reasoning, the prompts in Talk2Sensors are explicitly formulated based on heterogeneous physical traits. We categorize the instructions according to their required sensory cues. As shown in the distribution chart, 31.7\% of the queries rely exclusively on LiDAR geometric cues, while 20.7\% necessitate the joint fusion of LiDAR and radar. Crucially, instructions demanding cross-modal physical reasoning constitute a substantial portion of the dataset, for instance, the combination of camera and LiDAR accounts for 18.4\%, and queries requiring the comprehensive fusion of camera, LiDAR and radar account for 2.9\%. This diversified distribution forces perception models to dynamically route features across modalities rather than relying on a single dominant sensor.

\subsubsection{Spatial and Categorical Diversity}
To analyze the spatial distribution of the referred objects, we visualize the bird's-eye-view (BEV) location heatmaps for distinct categories (Fig. \ref{fig:dataset_stats_2}), including Pedestrians, Cars, Cyclists, and Trucks. The heatmaps demonstrate that vehicles (cars and trucks) are densely populated along the longitudinal driving lanes , whereas pedestrians and cyclists exhibit a broader lateral spread, accurately reflecting complex sidewalk and intersection dynamics. More importantly, by incorporating 4D radar and LiDAR, Talk2Sensors effectively captures objects at extended depth ranges and across varied kinematic states (e.g., moving vs. static), introducing novel challenges for robust, physics-aware scene understanding.

\section{Method}\label{sec:method}
The property-aware nature of Talk2Sensors directly shapes the design of TSFormer. Since each prompt refers to attributes that only particular sensors can perceive, texture from the camera, precise geometry from LiDAR, and motion from radar, the framework should retrieve evidence selectively from heterogeneous and unevenly distributed signals, rather than fusing all modalities indiscriminately. Accordingly, as Fig. \ref{fig:pipeline} shows, TSFormer is built as a unified, end-to-end architecture that grounds language in the appropriate physical properties through a coarse-to-fine, language-guided process.
Given a driving scene and a natural-language instruction, modality-specific backbones extract unimodal features $\mathbf{F}_{m}$, $m \in \{cam, lid, rad\}$, while a pre-trained text encoder produces a global semantic representation $F_{text} \in \mathbb{R}^{C}$. To relate these heterogeneous features without rigid, calibration-dependent projection, the Language-Routed Property Sampler (LRPS) recasts sampling as a query-driven process in a unified space: it modulates modality-specific sampling attention with query-level linguistic cues for coarse, language-conditioned retrieval, and then performs property-aligned deformable sampling to collect fine-grained cues $\mathbf{SF}_{m}$ from each sensor. The sampled cues are subsequently arbitrated by the Sparse-Preserving Modality Arbiter module, whose soft-gated sparse cross-attention emphasizes query-relevant modalities and keeps dense camera responses from overwhelming the sparse yet decisive radar signal. A shared transformer decoder finally refines the fused query representation $\mathbf{SF}_{fus}$ to predict the referred 3D bounding boxes.

\begin{figure}
    \includegraphics[width=0.99\linewidth]{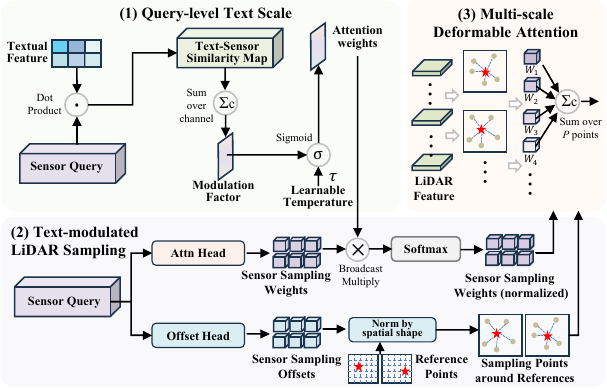}
    \vspace{-3mm}
    \caption{The structure of Language-Routed Property Sampler (LRPS). By injecting global text embeddings to initialize object queries and modulate deformable sampling offsets, LRPS efficiently extracts property-aligned preliminary features from multi-modal sensor maps.}
    \label{fig:lrps}
\end{figure}

\subsection{Language-Routed Property Sampler}

To bridge linguistic intent and sensor-specific physical cues, we propose a Language-Routed Property Sampler (LRPS) as Fig. \ref{fig:lrps} shows. Instead of directly reweighting raw sensor features, LRPS first measures the semantic compatibility between object queries and the textual prompt, and then injects the resulting query-level language scale into the sampling attention weights. In this way, the sampler can guide each query to retrieve sensor features from text-relevant spatial regions.

Let $\mathbf{Q}\in\mathbb{R}^{B\times N_q\times C}$ denote the object queries, where $B$, $N_q$, and $C$ are the batch size, number of queries, and embedding dimension, respectively. Given the global textual feature $\mathbf{t}\in\mathbb{R}^{B\times C}$, we project it into the sensor-specific semantic space:
\begin{equation}
\mathbf{t}_{m}
=
\phi_m(\mathbf{t}),
\
\mathbf{t}_{m}\in\mathbb{R}^{B\times C},
\end{equation}
where $m$ denotes the sensor modality and $\phi_m(\cdot)$ is a modality-specific text projection. To compute the query-level text-sensor similarity, the projected text feature is broadcast along the query dimension and matched with object queries by channel-wise interaction:
\begin{equation}
\mathbf{D}_{m}
=
\sum_{c=1}^{C}
\left(
\mathbf{Q}_{:,:,c}
\odot
\mathbf{t}_{m,:,c}
\right),
\
\mathbf{D}_{m}\in\mathbb{R}^{B\times N_q}.
\end{equation}

Here, $\odot$ denotes element-wise multiplication with broadcasting. The similarity map $\mathbf{D}_{m}$ is further converted into a bounded modulation factor through a learnable temperature parameter:
\begin{equation}
\mathbf{G}_{m}
=
\sigma
\left(
\tau_m\mathbf{D}_{m}
\right),
\
\mathbf{G}_{m}\in\mathbb{R}^{B\times N_q},
\end{equation}
where $\sigma(\cdot)$ is the sigmoid function and $\tau_m$ is a learnable modality-specific temperature parameter initialized to one.

Taking the LiDAR branch as an example, the object queries generate raw sampling attention weights and sampling offsets through two lightweight heads:
\begin{equation}
\mathbf{A}_{m}
\in
\mathbb{R}^{B\times N_q\times H\times (LP)},
\
\Delta_m
\in
\mathbb{R}^{B\times N_q\times H\times L\times P\times 2},
\end{equation}
where $H$, $L$, and $P$ denote the number of attention heads, feature levels, and sampling points, respectively. The query-level modulation factor $\mathbf{G}_{m}$ is broadcast to the attention layout and applied before normalization:
\begin{equation}
\widetilde{\mathbf{A}}_{m}
=
\operatorname{\mathtt{Softmax}}_{LP}
\left(
\mathbf{A}_{m}
\odot
\mathcal{E}_{H,LP}(\mathbf{G}_{m})
\right),
\
\widetilde{\mathbf{A}}_{m}
\in
\mathbb{R}^{B\times N_q\times H\times L\times P},
\end{equation}
where $\mathcal{E}_{H,LP}(\cdot)$ denotes broadcasting to the head and sampling dimensions. Meanwhile, the sampling locations are obtained by adding normalized offsets to the reference points:
\begin{equation}
\mathbf{S}_{m}
=
\mathbf{R}_{m}
+
\frac{\Delta_m}{\boldsymbol{\Omega}_{m}},
\
\mathbf{S}_{m}
\in
\mathbb{R}^{B\times N_q\times H\times L\times P\times 2},
\end{equation}
where $\mathbf{R}_{m}$ represents the reference points and $\boldsymbol{\Omega}_{m}$ is the spatial normalizer. Finally, multi-scale deformable attention aggregates sensor values around the sampled locations:
\begin{equation}
\mathbf{O}_{m}
=
\operatorname{\mathtt{MSDeformAttn}}
\left(
\mathbf{V}_{m},
\mathbf{S}_{m},
\widetilde{\mathbf{A}}_{m}
\right),
\
\mathbf{O}_{m}\in\mathbb{R}^{B\times N_q\times C}.
\end{equation}

This design makes the sampling process explicitly language-aware: the text prompt first produces a query-level modulation factor, which then adjusts the sensor sampling weights before deformable aggregation. As a result, LRPS can retrieve preliminary sensor features that are more consistent with the physical attributes described by the language query, such as geometry from LiDAR, appearance from cameras, and motion cues from radar.

\begin{figure}
    \includegraphics[width=0.99\linewidth]{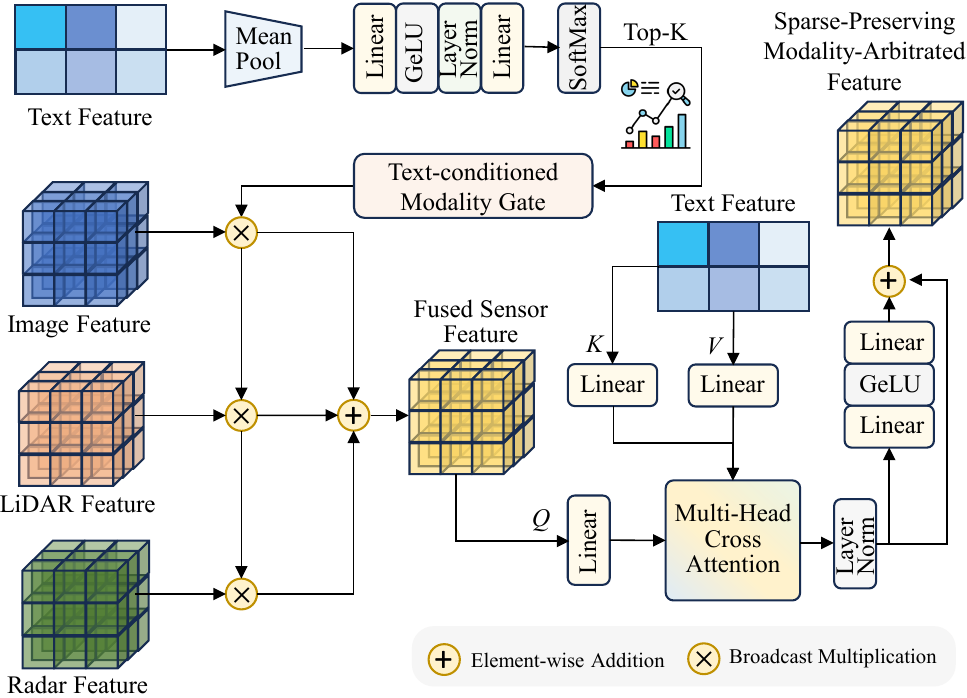}
    \vspace{-3mm}
    \caption{The structure of the Sparse-Preserving Modality Arbiter. A text-conditioned gate fuses the sampled sensor features while preventing dense modalities from overwhelming sparse but critical signals, and a text-guided cross-attention (fused feature as query, text tokens as keys/values) refines the result into $\mathbf{F}_{\mathrm{SPMA}}$.}
    \label{fig:sparse_attention}
\end{figure}

\subsection{Sparse-Preserving Modality Arbiter}

The Language-Routed Property Sampler provides a coarse language-conditioned anchoring by retrieving preliminary sensor features around query-relevant regions. However, these sampled features may still contain background responses and modality-specific redundancy, especially in complex driving scenes where different sensors capture heterogeneous and unevenly distributed physical cues. Therefore, after coarse property-aware sampling, we introduce a Sparse-Preserving Modality Arbiter module to perform fine-grained sensor arbitration and textual refinement, aiming to determine the precise referred spatial location indicated by the language prompt.

Specifically, this module first predicts text-conditioned modality gates from the global language representation (Fig. \ref{fig:sparse_attention}), adaptively selecting the most relevant sensor cues for each input sample. The gated sensor features are then aggregated into a unified fused sensor representation, which is further aligned with the text tokens through a final multi-head cross-attention layer. In this way, the model progressively transitions from coarse language-guided spatial anchoring to fine-grained cross-modal reasoning for accurate 3D visual grounding.

Let $\mathbf{O}_m\in\mathbb{R}^{B\times N_q\times C}$ denote the sampled feature of modality $m\in\mathcal{M}$, where $\mathcal{M}\subseteq\{\mathrm{cam},\mathrm{lidar},\mathrm{radar}\}$ is the set of available modalities. Let $\mathbf{T}\in\mathbb{R}^{B\times L_t\times C}$ denote the text token features. We first summarize the textual sequence into a global semantic vector: \begin{equation} 
\mathbf{t}_g = \phi_g \left( \frac{1}{L_t} \sum_{\ell=1}^{L_t} \mathbf{T}_{:,\ell,:} \right), \ \mathbf{t}_g\in\mathbb{R}^{B\times C}, \end{equation} 
where $\phi_g(\cdot)$ denotes a learnable projection layer. The global text vector is then fed into a lightweight modality gate network: \begin{equation} 
\mathbf{z} = \eta_g(\mathbf{t}_g), \ \mathbf{z}\in\mathbb{R}^{B\times M}, 
\end{equation} where $M=|\mathcal{M}|$, and $\eta_g(\cdot)$ is implemented by a linear layer, a non-linear activation, layer normalization, and an output projection. To encourage sparse modality routing, an optional Top-$K$ selection mask $\mathbf{m}_{\mathrm{topk}}\in\{0,1\}^{B\times M}$ is applied to the gate logits: 
\begin{equation} \mathbf{z}^{\prime} = \mathbf{z} \odot \mathbf{m}_{\mathrm{topk}}. 
\end{equation} 
When Top-$K$ selection is disabled, we set $\mathbf{z}^{\prime}=\mathbf{z}$. The text-conditioned modality gate is obtained by 
\begin{equation} 
\boldsymbol{\alpha} = \operatorname{\mathtt{Softmax}}_{M} \left( \mathbf{z}^{\prime} \right), \ \boldsymbol{\alpha} = [ \alpha_{\mathrm{cam}}, \alpha_{\mathrm{lidar}}, \alpha_{\mathrm{radar}} ] \in \mathbb{R}^{B\times M}. \end{equation} 
Since $\boldsymbol{\alpha}$ is predicted from the global text representation, it acts as a sample-level modality gate rather than a per-query gate. Each scalar gate is broadcast to all query tokens and channels, and the modality-specific features are fused by weighted summation: 
\begin{equation} \mathbf{F}_{\mathrm{s}} = \sum_{m\in\mathcal{M}} \alpha_m \mathbf{O}_m, \ \mathbf{F}_{\mathrm{s}} \in \mathbb{R}^{B\times N_q\times C}. 
\end{equation} 
This step performs semantic-driven sensor arbitration, allowing the network to emphasize the modalities that are most consistent with the textual requirement. To further align the fused sensor tokens with the fine-grained linguistic intent, we employ a final text-guided multi-head cross-attention layer. Specifically, the fused sensor feature is used as the query, while the text tokens serve as keys and values: \begin{equation} 
\mathbf{Q}_{s} = \mathbf{F}_{\mathrm{s}}\mathbf{W}_{Q}, \ \mathbf{K}_{t} = \mathbf{T}\mathbf{W}_{K}, \ \mathbf{V}_{t} = \mathbf{T}\mathbf{W}_{V}. \end{equation} 

For each attention head, the text-guided sensor representation is computed as 
\begin{equation} \operatorname{\mathtt{Attn}} \left( \mathbf{Q}_{s}, \mathbf{K}_{t}, \mathbf{V}_{t} \right) = \operatorname{\mathtt{Softmax}} \left( \frac{ \mathbf{Q}_{s}\mathbf{K}_{t}^{\top} }{ \sqrt{d_h} } \right) \mathbf{V}_{t}, 
\end{equation} 
where $d_h$ is the channel dimension of each head. The outputs of all heads are concatenated and linearly projected: 
\begin{equation} \mathbf{F}_{\mathrm{c}} = \operatorname{\mathtt{MHA}} \left( Q=\mathbf{F}_{\mathrm{s}}, K=\mathbf{T}, V=\mathbf{T} \right), \ \mathbf{F}_{\mathrm{c}}\in\mathbb{R}^{B\times N_q\times C}. 
\end{equation} 

Finally, the cross-attended feature is refined by a feed-forward network with residual connection and layer normalization: 
\begin{equation} \mathbf{Y} = \operatorname{\mathtt{LN}} \left( \mathbf{F}_{\mathrm{c}} \right), \end{equation} \begin{equation} \mathbf{F}_{\mathrm{SPMA}} = \operatorname{\mathtt{LN}} \left( \mathbf{Y} + \operatorname{\mathtt{FFN}} \left( \mathbf{Y} \right) \right), \ \mathbf{F}_{\mathrm{SPMA}}\in\mathbb{R}^{B\times N_q\times C}. 
\end{equation} 

Here, $\mathbf{F}_{\mathrm{SPMA}}$ denotes the final sparse-preserving modality-arbitrated feature. It integrates text-conditioned modality selection, gated sensor aggregation, and text-guided cross-attention refinement, providing a compact query-level representation for subsequent decoding and grounding prediction.

\subsection{Grounding Decoder and Prediction Head}

After sparse-preserving modality arbitration, the output feature 
$\mathbf{F}_{\mathrm{SPMA}}\in\mathbb{R}^{B\times N_q\times C}$ 
serves as the language-conditioned query representation for final 3D grounding. Instead of introducing task-specific post-processing, we adopt a query-based set prediction decoder to refine these features and produce a fixed-size set of grounding candidates. This design provides a clean prediction interface, allowing the proposed language-guided sampling and fusion modules to be evaluated without being coupled with heuristic proposal generation or hand-crafted matching rules.

Let $\mathbf{H}^{0}=\mathbf{F}_{\mathrm{SPMA}}$ denote the initial query features fed into the grounding decoder. A stack of decoder layers progressively refines the query representations:
\begin{equation}
\mathbf{H}^{\ell}
=
\mathcal{D}_{\ell}
\left(
\mathbf{H}^{\ell-1}
\right),
\
\ell=1,\ldots,L_d,
\end{equation}
where $\mathcal{D}_{\ell}(\cdot)$ denotes the $\ell$-th decoder layer and $L_d$ is the number of decoder layers. The final-layer representation 
$\mathbf{H}^{L_d}\in\mathbb{R}^{B\times N_q\times C}$ 
is used for grounding prediction. Each query token is independently decoded by two lightweight branches: a referring score branch and a 3D box regression branch. We emphasize that the referring branch does not predict semantic categories; instead, it outputs a single binary confidence $\hat{\alpha}_i\in[0,1]$ indicating whether the $i$-th query corresponds to the object referred by the language prompt. Object semantics are conveyed entirely through the prompt, so the model performs language-conditioned localization rather than category-level detection, consistent with the visual grounding formulation~\cite{chen2020scanrefer, achlioptas2020referit3d}.
\begin{equation}
\hat{\alpha}_i
=
\sigma
\left(
f_{\mathrm{cls}}
\left(
\mathbf{h}^{L_d}_i
\right)
\right),
\end{equation}
\begin{equation}
\hat{\mathbf{s}}_i
=
f_{\mathrm{box}}
\left(
\mathbf{h}^{L_d}_i
\right),
\end{equation}
where $\mathbf{h}^{L_d}_i\in\mathbb{R}^{C}$ is the feature of the $i$-th query, $\hat{\alpha}_i\in[0,1]$ denotes the referring confidence (a prompt-conditioned matching score), and $\hat{\mathbf{s}}_i$ denotes the predicted 3D box. Specifically, each prediction is represented as

\begin{equation}
\hat{\mathbf{b}}_i
=
\left(
\hat{\alpha}_i,
\hat{\mathbf{s}}_i
\right),
\
i=1,\ldots,N_q,
\end{equation}
with
\begin{equation}
\hat{\mathbf{s}}_i
=
\left(
\hat{x}_i,
\hat{y}_i,
\hat{z}_i,
\hat{w}_i,
\hat{h}_i,
\hat{l}_i,
\hat{\theta}_i
\right).
\end{equation}

Here, $(\hat{x}_i,\hat{y}_i,\hat{z}_i)$ denotes the 3D center, $(\hat{w}_i,\hat{h}_i,\hat{l}_i)$ denotes the spatial size, and $\hat{\theta}_i$ denotes the yaw angle. Following common practice in 3D detection, the yaw angle is optimized with a sine-cosine representation to avoid angular discontinuity.

The decoder predicts a set of candidates 
$\hat{\mathcal{B}}=\{\hat{\mathbf{b}}_i\}_{i=1}^{N_q}$, 
where only a small subset is expected to correspond to the referred objects described by the prompt. Therefore, the grounding head does not rely on non-maximum suppression or manually designed proposal selection. Instead, the predicted set is supervised through bipartite matching, as described in the training objective. In this sense, the prediction head acts as a standardized grounding readout, while the core contribution of TSFormer lies in producing more discriminative language-conditioned query features through property-aware sampling and sparse-preserving modality arbitration.

\subsection{Training Objectives}
\label{sec:training_objectives}
TSFormer formulates 3D visual grounding as a set prediction problem following the DETR paradigm~\cite{carion2020end}. Given $N_q$ object queries, the transformer decoder produces a set of predictions $\hat{\mathcal{B}} = \{\hat{b}_i\}_{i=1}^{N_q}$. Each prediction $\hat{b}_i = (\hat{\alpha}_i, \hat{s}_i)$ consists of a \textit{referring confidence score} $\hat{\alpha}_i \in [0, 1]$, which indicates the likelihood that the query matches the linguistic description, and a 3D bounding box $\hat{s}_i = (\hat{x}_i, \hat{y}_i, \hat{z}_i, \hat{w}_i, \hat{h}_i, \hat{l}_i, \hat{\theta}_i)$. This bounding box is parameterized by its center coordinates $(x, y, z)$, spatial dimensions (width $w$, height $h$, length $l$), and yaw orientation $\theta$. Following standard practices in 3D object detection for autonomous driving~\cite{lang2019pointpillars, yan2018second}, we encode the yaw angle $\theta$ using a sine-cosine representation, $\theta_{\text{enc}} = [\sin\theta, \cos\theta]$, to resolve the $2\pi$-periodicity discontinuity prior to applying L1 regression on the encoded values.Since the number of referred objects $M$ varies per scene and typically $M \ll N_q$, we establish a correspondence between the predictions and the ground truth objects via optimal bipartite matching. Let $\mathcal{G} = \{g_j\}_{j=1}^{M}$ denote the ground truth set, where each $g_j = (\alpha_j, s_j)$ with $\alpha_j = 1$ indicating a referred object and $s_j$ its 3D bounding box. We seek the optimal assignment $\sigma^* \in \mathfrak{S}_{N_q}$ that minimizes the total matching cost:
\begin{equation}
\sigma^* = \arg\min_{\sigma \in \mathfrak{S}_{N_q}} \sum_{j=1}^{M} \mathcal{C}_{\text{match}}\bigl(\hat{b}_{\sigma(j)}, g_j\bigr),
\end{equation}
where the pairwise matching cost $\mathcal{C}_{\text{match}}$ integrates the referring confidence and 3D box similarity:
\begin{equation}
\begin{aligned}
    \mathcal{C}_{\text{match}}\bigl(\hat{b}_{\sigma(j)}, g_j\bigr) = & -\hat{\alpha}_{\sigma(j)} + \lambda_{\text{L1}} \bigl\|\hat{s}_{\sigma(j)} - s_j\bigr\|_1 \\
& + \lambda_{\text{IoU}} \mathcal{L}_{\text{GIoU}}^{3D}\bigl(\hat{s}_{\sigma(j)}, s_j\bigr).
\end{aligned}
\label{eq:matching_cost}
\end{equation}
where the first item encourages high-confidence predictions for referred objects; the second term penalizes the L1 distance between predicted and ground-truth box parameters; the third term computes the 3D Generalized IoU (GIoU)~\cite{rezatofighi2019generalized} between two 3D oriented boxes, providing scale-invariant geometric supervision. The optimal assignment $\sigma^*$ is computed efficiently via the Hungarian algorithm~\cite{kuhn1955hungarian}.

Following the standard practice in 3D object detection for autonomous driving~\cite{lang2019pointpillars, yan2018second}, we encode the yaw angle $\theta$ using the sine-cosine representation $\theta_{\text{enc}} = [\sin\theta, \cos\theta]$ to resolve the $2\pi$-periodicity discontinuity, and apply L1 regression on the encoded values.

Once the optimal assignment $\sigma^*$ is established, the grounding loss for the final decoder layer is computed over all matched pairs as well as unmatched predictions:
\begin{equation}
\begin{aligned}
\mathcal{L}_{\text{ground}} = & \sum_{j=1}^{M} \Bigl[ \mathcal{L}_{\text{ce}}\bigl(\hat{\alpha}_{\sigma^*(j)}, 1\bigr) + \lambda_{\text{L1}} \bigl\|\hat{s}_{\sigma^*(j)} - s_j\bigr\|_1 \\ & + \lambda_{\text{IoU}} \mathcal{L}_{\text{GIoU}}^{3D}\bigl(\hat{s}_{\sigma^*(j)}, s_j\bigr) \Bigr] \\
& + \sum_{i \in \Omega_{\text{unmtch}}} \mathcal{L}_{\text{ce}}\bigl(\hat{\alpha}_i, 0\bigr),
\end{aligned}
\label{eq:ground_loss}
\end{equation}
where $\mathcal{L}_{\text{ce}}(\cdot, \cdot)$ denotes the cross-entropy loss for referring confidence estimation; $\Omega_{\text{unmtch}}$ indexes all unmatched predictions, which are supervised as negatives. Box regression is only applied to matched foreground objects.

To facilitate gradient propagation through the deep transformer decoder and accelerate convergence, we impose auxiliary supervision at each intermediate decoding layer. Concretely, each decoder layer $l \in \{1, \dots, L-1\}$ independently produces predictions $\hat{\mathcal{B}}^{(l)} = \{\hat{b}_i^{(l)}\}_{i=1}^{N_q}$, which are matched against the same ground truth set $\mathcal{G}$ via separate Hungarian matching. The auxiliary loss aggregates the grounding loss across all intermediate layers:
\begin{equation}
\mathcal{L}_{\text{aux}} = \sum_{l=1}^{L-1} \mathcal{L}_{\text{ground}}^{(l)},
\end{equation}
where $\mathcal{L}_{\text{ground}}^{(l)}$ shares the identical formulation as Eq.~\eqref{eq:ground_loss} but operates on the $l$-th layer's outputs. This progressive intermediate supervision guides the model to iteratively refine query representations before the final prediction.

Finally, the complete training objective combines the final-layer grounding loss with all auxiliary decoding losses:
\begin{equation}
\mathcal{L}_{\text{total}} = \mathcal{L}_{\text{ground}}^{(L)} + \lambda_{\text{aux}} \cdot \mathcal{L}_{\text{aux}}.
\label{eq:total_loss}
\end{equation}

\begin{table*}
    \centering
    \caption{The overall performances of models on Talk2Sensors dataset. \textbf{Bold} denotes the best performance of the current metric.}
    \vspace{-3mm}
    \setlength\tabcolsep{2.9pt}
    \begin{tabular}{cc|cc|cccc|c|cccc|c}
    \toprule
       \multirow{2}[2]{*}{\textbf{Models}} & \multirow{2}[2]{*}{\textbf{Venues}} & \multirow{2}[2]{*}{\textbf{Sensors}} & \multirow{2}[2]{*}{\textbf{Text Encoder}} & \multicolumn{5}{c}{\textbf{Entire Annotated Area (EAA)}}  & \multicolumn{5}{c}{\textbf{Driving Corridor Area (DCA)}}  \\
       \cmidrule(lr){5-9}  \cmidrule(lr){10-14}
        & & & & \textbf{Car} & \textbf{Pedestrian} & \textbf{Cyclist} & \textbf{mAP} & \textbf{mAOS} & \textbf{Car} & \textbf{Pedestrian} & \textbf{Cyclist} & \textbf{mAP} & \textbf{mAOS}  \\
    \midrule
    \textbf{TSFormer (ours)} & 2026 & C + L + R5 & No Text & 0.014 &	0.000 &	0.006 &	0.007 &	0.059 &	0.033 &	0.000 &	0.015 &	0.016 &	0.059 \\
    \midrule
    CenterPoint & CVPR$_{2021}$ & R5 & CLIP & 0.082 & 0.202 & 0.128 & 0.137 & 0.009 & 0.278 & 0.253 & 0.138 & 0.223 & 0.098 \\
    T-RadarNet & ICRA$_{2025}$ & R5 & CLIP & 0.147 & 0.505 & 0.138 & 0.263 & 0.103 & 0.241 & 0.505 & 0.826 & 0.524 & 0.318 \\
    \textbf{TSFormer (ours)} & 2026 & R5 & CLIP & 1.433 & 1.909 & 0.146 & 1.463 & 1.217 & 1.769 & 1.010 & 1.127 & 1.635 & 1.396 \\  
    \midrule
    BEVFormer & ECCV$_{2022}$ & C & CLIP & 9.360 & 4.638 & 7.687 & 7.228 & 5.196 & 11.397 & 4.838 & 13.747 & 9.994 & 8.016 \\
    Mono3DVG-TR & AAAI$_{2024}$ & C & CLIP & 10.413 & 0.832 & 3.780 & 5.008 & 4.143 & 15.448 & 0.828 & 3.126 & 6.468 & 4.557 \\
    \textbf{TSFormer (ours)} & 2026 & C & CLIP & 12.667 & 5.004 & 12.629 & 10.100 & 20.322 & 32.979 & 5.176 & 10.850 & 16.335 & 25.176 \\
    \midrule
    FUTR3D & CVPRW$_{2023}$ & C + R5 & CLIP & 9.451 & 3.309 & 10.110 & 7.623 & 5.032 & 12.577 & 3.459 & 15.970 & 10.669 & 7.879 \\
    BEVFusion & ICRA$_{2023}$ & C + R5 & CLIP & 12.010 & 10.484 & 12.308 & 11.601 & 8.934 & 18.732 & 11.195 & 17.457 & 15.794 & 13.072 \\
    \textbf{TSFormer (ours)} & 2026 & C + R5 & CLIP & 10.928 & 12.136 & 16.360 & 13.141 & 18.460 & 16.701 & 14.034 & 24.141 & 18.292 & 27.737 \\
    \midrule
    Ges3ViG & CVPR$_{2025}$ & L & CLIP & 51.178 & 41.244 & 33.355 & 41.926 & 38.869 & 78.265 & 49.732 & 43.121 & 57.040 & 52.429 \\
    TSP3D & CVPR$_{2025}$ & L & CLIP & 51.793 & 41.579 & 30.322 & 41.231 & 38.897 & 86.120 & 48.253 & 34.546 & 56.306 & 51.698 \\
    TPCNet & TITS$_{2026}$ & L & CLIP & 50.794 & 41.585 & 34.161 & 42.180 & 37.762 & 77.425 & 48.771 & 43.024 & 56.407 & 52.732 \\ 
    \textbf{TSFormer (ours)} & 2026 & L & CLIP & 56.123 & 43.647 & 38.776 & 46.182 & 40.412 & 84.294 & 50.781 & 44.231 & 59.768 & 53.343 \\
    \midrule
    TPCNet & T-ITS$_{2026}$ & L + R5 & CLIP & 50.747 & 43.198 & 39.461 & 44.469 & 42.036 & 77.273 & 52.412 & 47.506 & 59.064 & 56.757 \\
    \textbf{TSFormer (ours)} & 2026 & L + R5 & CLIP & 55.029 & 46.633 & \textbf{43.164} & 48.275 & 41.044 & 84.444 & 54.225 & 51.614 & 63.428 & 57.355 \\
    \midrule
    FUTR3D & CVPRW$_{2023}$ & C + L + R5 & CLIP & 51.162 & 43.828 & 33.857 & 42.949 & 39.017 & 78.984 & 51.600 & 46.051 & 58.878 & 54.496 \\
    \rowcolor{gray!30} \textbf{TSFormer (ours)} & 2026 & C + L + R5 & CLIP & \textbf{67.558} & \textbf{47.462} & 37.981 & \textbf{51.000} & \textbf{48.437} & \textbf{89.033} & \textbf{56.758} & \textbf{53.707} & \textbf{66.499} & \textbf{63.216} \\
    \bottomrule
    \end{tabular}
    
    \label{tab:experiment_benchmark}
\end{table*}

\section{Experiments}\label{sec:experiments}

\subsection{Datasets and Evaluation Metrics}
To evaluate models across diverse sensor configurations, we conduct experiments on two 3D visual grounding datasets:
\textbf{1) Talk2Sensors:} As the first tri-modal benchmark, it strictly synchronizes Camera, LiDAR, and 4D mmWave Radar data. Its prompts emphasize objective physical measurements, requiring models to perform complex cross-modal reasoning over fine-grained visual textures, 3D geometry, and absolute kinematic states (e.g., radial velocity). Talk2Sensors comprises 8,682 referring prompts with 20,558 objects.
\textbf{2) Mono3DRefer:} Built upon KITTI, Mono3DRefer~\cite{zhan2024mono3dvg} is a 3D visual grounding benchmark that localizes referred objects from a single RGB image. It provides 41,140 ChatGPT-generated and manually refined expressions over 8,228 objects across 2,025 scenes, with descriptions jointly encoding appearance (e.g., color, category) and geometry (e.g., height/length, distance, azimuth) cues. Although the benchmark itself is defined on monocular RGB, its underlying KITTI source additionally offers synchronized LiDAR point clouds. We therefore evaluate all baselines under the standard monocular setting, while further supplementing the corresponding KITTI LiDAR, so as to assess the cross-dataset generalization of our property-aware design when a complementary geometric modality is available.

Following the View-of-Delft evaluation protocol~\cite{palffy2022multi}, we report results on two spatial regions: the Entire Annotated Area (EAA), spanning the full annotated range, and the Driving Corridor Area (DCA), the safety-critical corridor directly ahead of the ego-vehicle. For each region we report the mean Average Precision (mAP) over the Car, Pedestrian, and Cyclist classes, together with the mean Average Orientation Similarity (mAOS), which jointly measures localization and yaw-orientation accuracy.

\subsection{Implementation Details}
We benchmark TSFormer against a broad range of models. For general multi-sensor 3D detectors, we adopt FUTR3D~\cite{chen2023futr3d} and BEVFusion~\cite{liu2023bevfusion}. For dedicated 3D visual grounding, we compare with single-modal methods (T-RadarNet~\cite{guan2024talk2radar}, MSSG~\cite{cheng2023language}, AFMNet~\cite{solgi2024transformer}, EDA~\cite{wu2023eda}, Ges3ViG~\cite{mane2025ges3vig}, and TSP3D~\cite{guo2025text}) and the multi-modal TPCNet~\cite{guan2025talk2pc}. We further isolate our two modules: the Language-Routed Property Sampler is compared against MAFS~\cite{chen2023futr3d} and the DETR3D sampler~\cite{wang2022detr3d}, and the Sparse-Preserving Modality Arbiter against Sum, Concat-MLP, vanilla Cross-Attention, GMU~\cite{arevalo2017gated}, and SE-style gated fusion~\cite{hu2018squeeze}. For these module-level comparisons, only the corresponding component is swapped while all other settings remain fixed. Since the general detectors are not natively language-aware, we grant them identical guidance by injecting the CLIP text feature into their FPN-stage features via a vanilla cross-attention layer (sensor features as queries, text tokens as keys/values); all baselines are re-implemented under the same backbone, query setting, and training schedule for fairness.

TSFormer is built upon FUTR3D with a unified camera-LiDAR-radar-language architecture. Images are encoded by ResNet-101 with FPN; LiDAR points are voxelized at $0.05\times0.05\times0.1$~m and encoded by a SparseEncoder-SECOND backbone with an FPN neck; radar is processed by RadarFeatureNet and PointPillarsScatter at $0.2\times0.2\times0.1$~m; and text is encoded by a pretrained CLIP ViT-B/32 (max.\ 77 tokens). The DETR-style grounding decoder uses 900 object queries and six layers, with embedding dimension 256, feed-forward dimension 1024, and 8 heads. The modality-specific text projection in the sampler has a hidden dimension of 128, and the arbiter's routing network is two linear layers with intermediate LayerNorm and ReLU.

We train with AdamW (learning rate and weight decay both $1\times10^{-4}$) for 80 epochs, using a batch size of 2 per GPU on 4 NVIDIA RTX 4090 GPUs, gradient clipping at norm 35, and a RepeatDataset factor of 2; augmentation includes random flipping and range filtering. At inference, we filter detections at a confidence of 0.1 and apply circle-NMS (IoU 0.2), keeping up to 300 boxes within the range $[0, -25.6, -3, 51.2, 25.6, 2]$~m.

\begin{figure}
    \includegraphics[width=0.99\linewidth]{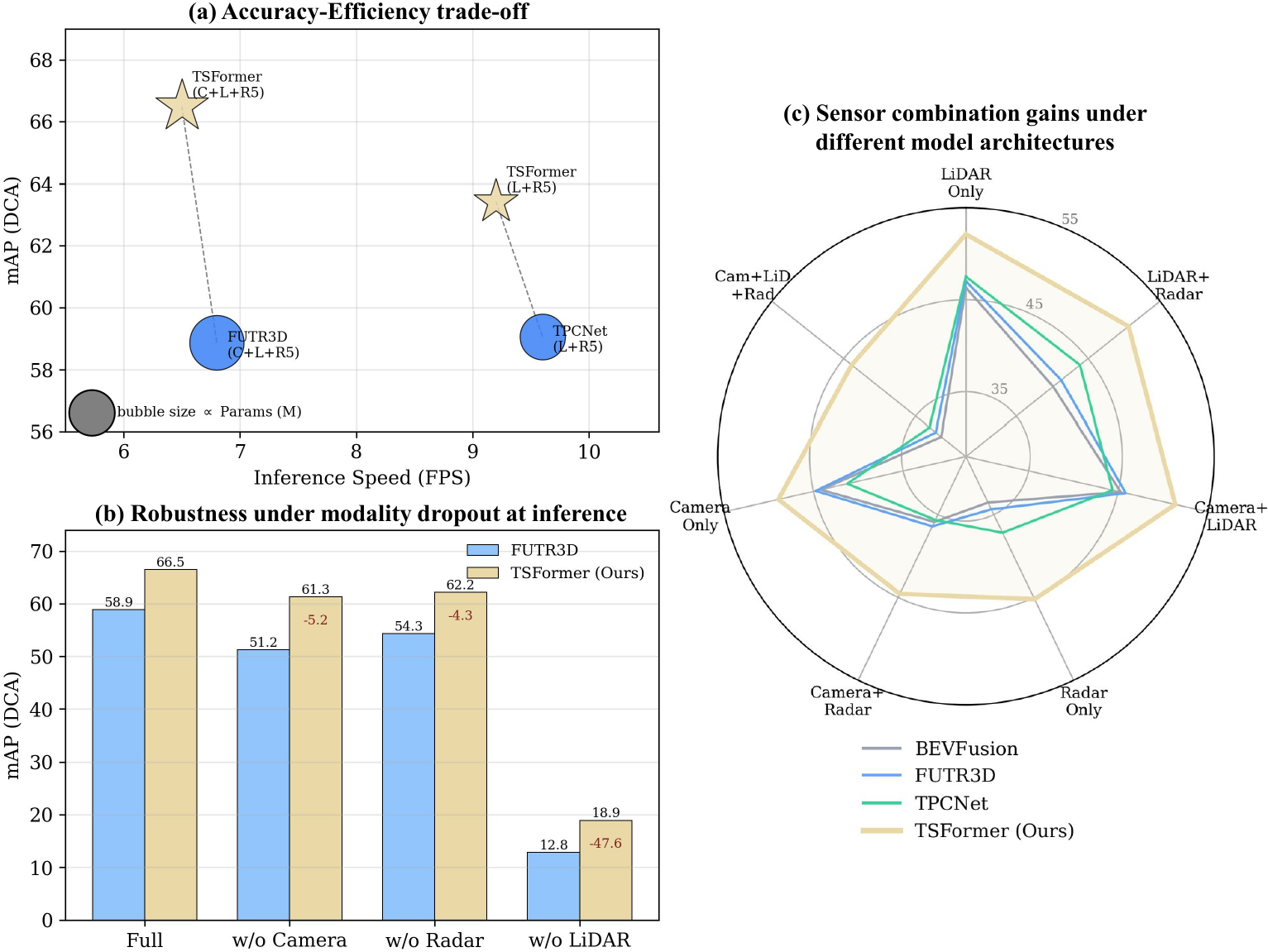}
    \vspace{-3mm}
    \caption{The comparison regarding efficiency, robustness and sensor combination gains. (a) Accuracy-Efficiency trade-off; (b) Robustness under modality dropout at inference; (c) Sensor combination gains under different model architectures.}
    \label{fig:stat_figures}
\end{figure}

\subsection{Quantitative Results}

\textbf{Overall benchmark.}
Table~\ref{tab:experiment_benchmark} reports the main comparison. TSFormer attains the best mAP/mAOS under every sensor configuration, and the full tri-sensor model reaches $51.000$/$66.499$ mAP on EAA/DCA, surpassing the strongest fusion baseline FUTR3D by $+8.05$/$+7.62$. Two observations are worth highlighting. First, the text-free variant collapses to near-zero mAP ($0.007$), confirming that the task is genuinely referential rather than exhaustive detection. Second, LiDAR geometry dominates single-modality grounding ($46.182$ mAP) while camera ($10.100$) and radar ($1.463$) are far weaker alone, yet naively fusing all three does not help FUTR3D (C+L+R5, $42.949$) beyond its LiDAR-only level, as dense camera tokens drown the sparse radar signal. In contrast, TSFormer improves monotonically as modalities are added ($46.18\!\rightarrow\!48.28\!\rightarrow\!51.00$ EAA mAP), evidencing that its language-guided routing actually extracts complementary value from each sensor. The orientation metric shows the same trend, with a large mAOS margin ($48.437$ vs.\ $39.017$), which matters for kinematics-oriented prompts.

\begin{table*}
\centering
\caption{Generalization comparison on the Mono3DRefer benchmark. Following Mono3DVG~\cite{zhan2024mono3dvg}, we report accuracy at 3D IoU thresholds of 0.25 and 0.5 on the \textbf{Unique}, \textbf{Multiple}, and \textbf{Overall} splits. \textbf{Bold} marks the best learning-based result.}
\vspace{-3mm}
\setlength\tabcolsep{10.0pt}
\label{tab:mono3drefer}
\setlength{\tabcolsep}{6pt}
\renewcommand{\arraystretch}{1.15}
\begin{tabular}{ccc| cc cc cc}
\toprule
\multirow{2}{*}{\textbf{Methods}} & \multirow{2}{*}{\textbf{Venues}} & \multirow{2}{*}{\textbf{Types}} & \multicolumn{2}{c}{\textbf{Unique}} & \multicolumn{2}{c}{\textbf{Multiple}} & \multicolumn{2}{c}{\textbf{Overall}} \\
\cmidrule(lr){4-5}\cmidrule(lr){6-7}\cmidrule(lr){8-9}
 & & & Acc@0.25 & Acc@0.5 & Acc@0.25 & Acc@0.5 & Acc@0.25 & Acc@0.5 \\
\midrule
ReSC\,+\,backproj         &  ECCV$_{2020}$  & One-Stage   & 11.96 & 0.49  & 23.69 & 3.94  & 21.48 & 3.29  \\
TransVG\,+\,backproj      &  ICCV$_{2021}$  & Tran.-based & 15.78 & 4.02  & 21.84 & 4.16  & 20.70 & 4.14  \\
Cube R-CNN          & CVPR$_{2023}$ & Two-Stage   & 35.29 & 16.67 & 60.52 & 32.99 & 55.77 & 29.92 \\
Mono3DVG-TR               &  AAAI$_{2024}$  & Tran.-based & 57.65 & 33.04 & \textbf{65.92} & 46.85 & \textbf{64.36} & 44.25\\
TPCNet  & T-ITS$_{2026}$ & One-Stage & 69.87 & 57.90 & 42.26 & 45.07 & 54.48 & 50.67    \\
\midrule
\textbf{TSFormer (Ours)} & 2026 & Tran.-based & \textbf{75.53} & \textbf{62.77} & 47.62 & \textbf{47.62} & 57.63 & \textbf{53.05} \\
\bottomrule
\end{tabular}
\end{table*}

\textbf{Where the gains come from.}
Grouping prompts by their required sensory cues (Fig.~\ref{fig:stat_figures} (c)) reveals that the improvement is not uniform: TSFormer's largest gains fall on radar-involved and tri-sensor prompts ($\Delta$Radar\,Only $+8.03$, Camera+Radar $+8.16$, Camera+LiDAR+Radar $+10.83$), precisely the settings where a single modality is insufficient and correct property routing is decisive. This directly substantiates the design motivation of the Language-Routed Property Sampler and Sparse-Preserving Modality Arbiter.

\textbf{Robustness and efficiency.}
Under inference-time modality dropout (Fig.~\ref{fig:stat_figures} (b)), TSFormer degrades more gracefully than FUTR3D when the camera ($-5.17$ vs.\ $-7.64$) or radar ($-4.34$ vs.\ $-4.56$) is removed, while both collapse without LiDAR, confirming LiDAR as the geometric backbone and the soft gate as an effective fallback for the auxiliary sensors. Meanwhile, the accuracy-efficiency trade-off (Fig.~\ref{fig:stat_figures} (a)) shows TSFormer dominating prior methods at comparable cost: it exceeds TPCNet by $+4.36$ mAP under L+R5 at similar parameters and near-identical speed ($9.2$ vs.\ $9.6$ FPS), retaining real-time-level throughput in the full tri-sensor setting.


\textbf{Cross-dataset generalization.}
Finally, on the monocular Mono3DRefer benchmark (Table~\ref{tab:mono3drefer}), TSFormer transfers without task-specific tuning and, leveraging the LiDAR available in KITTI, achieves the best accuracy at the strict $\text{IoU}\!=\!0.5$ threshold across Unique/Multiple/Overall splits ($53.05$ Overall vs.\ $44.25$ for the specialized Mono3DVG-TR), indicating more precise 3D localization and confirming the generality of the property-aware design.

\begin{figure*}
    \includegraphics[width=0.99\linewidth]{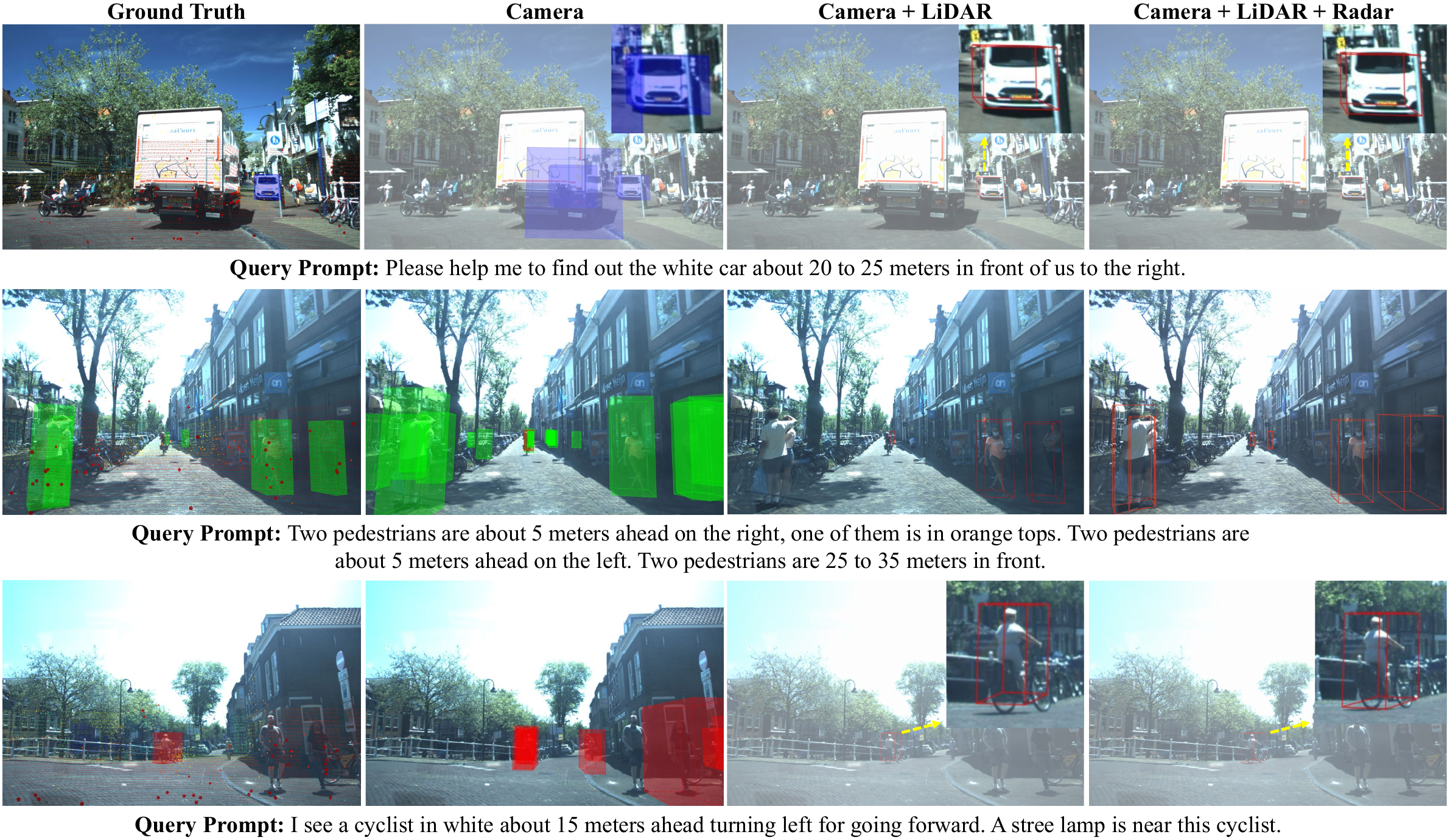}
    \vspace{-3mm}
    \caption{Qualitative grounding results of TSFormer under progressively enriched sensor configurations (Camera $\rightarrow$ Camera+LiDAR $\rightarrow$ Camera+LiDAR+Radar). }
    \label{fig:main_vis}
\end{figure*}

\subsection{Ablation Study}

\textbf{Component analysis.}
Tables~\ref{tab:ablation_main}-\ref{tab:spma_ablation} isolate each module. Starting from the modality-agnostic sampler with summation fusion ($43.217$ mAP), adding LRPS and SPMA contributes $+4.33$ and $+3.67$ independently and $+7.78$ jointly, indicating complementary rather than overlapping effects. LRPS outperforms the DETR3D sampler and MAFS ($51.000$ vs.\ $46.315$/$47.126$), and its query-level text modulation factor is the single most important internal component (removing it drops mAP to $47.230$). SPMA likewise beats all fusion alternatives including GMU and SE-style gating, with sample-level gating plus a light Top-$K$ ($K\!=\!2$) routing giving the best balance between sparsity and information retention.

\begin{table}
\centering
\caption{Ablation study of the proposed LRPS and SPMA on Talk2Sensors (C+L+R5). The baseline adopts the MAFS with simple summation fusion.}
\vspace{-3mm}
\setlength\tabcolsep{10.0pt}
\label{tab:ablation_main}
\begin{tabular}{cc|cc|cc}
\toprule
\multirow{2}{*}{\textbf{LRPS}} & \multirow{2}{*}{\textbf{SPMA}} & \multicolumn{2}{c|}{\textbf{EAA}} & \multicolumn{2}{c}{\textbf{DCA}} \\
\cmidrule(lr){3-4} \cmidrule(lr){5-6}
 & & \textbf{mAP} & \textbf{mAOS} & \textbf{mAP} & \textbf{mAOS} \\
\midrule
$\times$ & $\times$ & 43.217 & 39.528 & 59.104 & 54.836 \\
\checkmark & $\times$ & 47.542 & 44.263 & 62.847 & 59.032 \\
$\times$ & \checkmark & 46.885 & 43.617 & 62.138 & 58.341 \\
\rowcolor{gray!30} \checkmark & \checkmark & \textbf{51.000} & \textbf{48.437} & \textbf{66.499} & \textbf{63.216} \\
\bottomrule
\end{tabular}
\end{table}

\begin{table}
\centering
\caption{Comparison of different feature samplers on Talk2Sensors (C+L+R5). All variants share the identical SPMA module, decoder, and training schedule.}
\setlength\tabcolsep{8.0pt}
\vspace{-3mm}
\label{tab:sampler_comparison}
\begin{tabular}{l|cc|cc}
\toprule
\multirow{2}{*}{\textbf{Samplers}} & \multicolumn{2}{c|}{\textbf{EAA}} & \multicolumn{2}{c}{\textbf{DCA}} \\
\cmidrule(lr){2-3} \cmidrule(lr){4-5}
 & \textbf{mAP} & \textbf{mAOS} & \textbf{mAP} & \textbf{mAOS} \\
\midrule
DETR3D Sampler [28] & 46.315 & 43.128 & 61.542 & 57.913 \\
MAFS [21] & 47.126 & 43.845 & 62.437 & 58.652 \\
\rowcolor{gray!30} \textbf{LRPS (Ours)} & \textbf{51.000} & \textbf{48.437} & \textbf{66.499} & \textbf{63.216} \\
\bottomrule
\end{tabular}
\end{table}

\begin{table}
\centering
\caption{Comparison of different cross-modal fusion strategies on Talk2Sensors (C+L+R5). All variants share the identical LRPS module; only the fusion module is replaced.}
\vspace{-3mm}
\label{tab:fusion_comparison}
\begin{tabular}{l|cc|cc}
\toprule
\multirow{2}{*}{\textbf{Fusion Strategies}} & \multicolumn{2}{c|}{\textbf{EAA}} & \multicolumn{2}{c}{\textbf{DCA}} \\
\cmidrule(lr){2-3} \cmidrule(lr){4-5}
 & \textbf{mAP} & \textbf{mAOS} & \textbf{mAP} & \textbf{mAOS} \\
\midrule
Sum Fusion & 47.542 & 44.263 & 62.847 & 59.032 \\
Concat-MLP Fusion & 47.913 & 44.652 & 63.156 & 59.487 \\
Cross-Attention Fusion & 48.826 & 45.731 & 64.215 & 60.694 \\
GMU [29] & 49.317 & 46.108 & 64.823 & 61.245 \\
SE-style Gated Fusion [30] & 49.058 & 45.926 & 64.531 & 60.917 \\
\rowcolor{gray!30} \textbf{SPMA (Ours)} & \textbf{51.000} & \textbf{48.437} & \textbf{66.499} & \textbf{63.216} \\
\bottomrule
\end{tabular}
\end{table}

\begin{table}
\centering
\caption{Ablation on the internal designs of LRPS on Talk2Sensors (C+L+R5). Mod.\ Factor denotes the query-level text modulation factor $\mathbf{G}_m$; $\phi_m$ denotes the modality-specific text projection; $\tau_m$ denotes the learnable temperature.}
\vspace{-3mm}
\label{tab:lrps_ablation}
\resizebox{\columnwidth}{!}{
\begin{tabular}{l|cc|cc}
\toprule
\multirow{2}{*}{\textbf{Variant}} & \multicolumn{2}{c|}{\textbf{EAA}} & \multicolumn{2}{c}{\textbf{DCA}} \\
\cmidrule(lr){2-3} \cmidrule(lr){4-5}
 & \textbf{mAP} & \textbf{mAOS} & \textbf{mAP} & \textbf{mAOS} \\
\midrule
w/o Mod.\ Factor (plain deformable) & 47.230 & 44.019 & 62.518 & 58.746 \\
Shared text projection ($\phi_m \!\rightarrow\! \phi$) & 49.462 & 46.573 & 64.851 & 61.408 \\
Fixed temperature ($\tau_m \!=\! 1$) & 50.128 & 47.365 & 65.472 & 62.153 \\
Modulation after Softmax & 49.874 & 47.021 & 65.116 & 61.827 \\
\rowcolor{gray!30} \textbf{Full LRPS (Ours)} & \textbf{51.000} & \textbf{48.437} & \textbf{66.499} & \textbf{63.216} \\
\bottomrule
\end{tabular}}
\end{table}

\begin{table}
\centering
\caption{Ablation on the internal designs of SPMA on Talk2Sensors (C+L+R5). Text CA denotes the final text-guided cross-attention layer.}
\vspace{-3mm}
\label{tab:spma_ablation}
\resizebox{\columnwidth}{!}{
\begin{tabular}{l|cc|cc}
\toprule
\multirow{2}{*}{\textbf{Variant}} & \multicolumn{2}{c|}{\textbf{EAA}} & \multicolumn{2}{c}{\textbf{DCA}} \\
\cmidrule(lr){2-3} \cmidrule(lr){4-5}
 & \textbf{mAP} & \textbf{mAOS} & \textbf{mAP} & \textbf{mAOS} \\
\midrule
w/o modality gate (uniform weights) & 48.826 & 45.731 & 64.215 & 60.694 \\
Per-query gate (vs.\ sample-level) & 50.317 & 47.582 & 65.703 & 62.361 \\
w/o Text CA & 49.635 & 46.814 & 64.982 & 61.573 \\
Top-$K$ selection, $K\!=\!1$ & 49.216 & 46.335 & 64.517 & 61.089 \\
\rowcolor{gray!30} Top-$K$ selection, $K\!=\!2$ & \textbf{51.000} & \textbf{48.437} & \textbf{66.499} & \textbf{63.216} \\
Soft gating only (Top-$K$ disabled) & 50.628 & 47.953 & 66.084 & 62.735 \\
\bottomrule
\end{tabular}}
\end{table}


\subsection{Qualitative Analysis}
We further provide qualitative evidence that TSFormer grounds language in the correct physical properties.

\textbf{Progressive multi-sensor grounding.}
Fig.~\ref{fig:main_vis} visualizes predictions under progressively enriched sensor inputs. With the camera alone, the model can only anchor a coarse region on the image plane; lacking depth, it yields spatially inflated or displaced boxes and struggles once the prompt requires a metric distance (e.g., ``about 20 to 25 meters in front''). Introducing LiDAR restores accurate geometry, tightening the 3D extent and resolving depth-dependent references. Adding radar further supplies absolute kinematics: for motion-conditioned prompts such as ``a cyclist $\ldots$ turning left for going forward'', the radar velocity cues (yellow arrows) disambiguate a referent that appearance and geometry alone leave ambiguous. This progression mirrors the quantitative trend in Table~\ref{tab:experiment_benchmark}, confirming that each modality contributes complementary rather than redundant evidence.

\begin{figure}
    \includegraphics[width=0.998\linewidth]{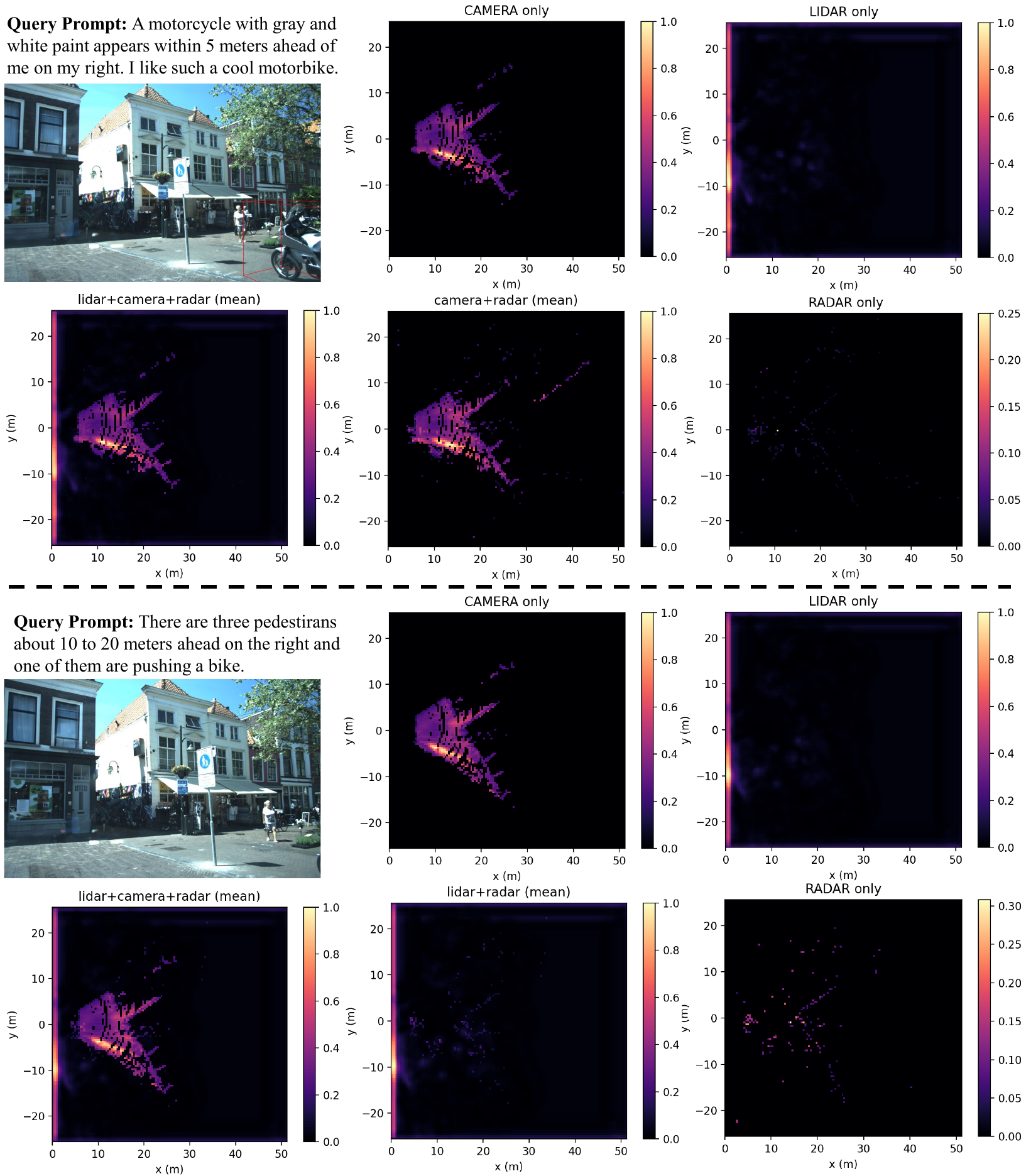}
    \vspace{-6mm}
    \caption{Heatmap visualization by various sensor configurations. For certain properties perceptible only through vision (such as color), the heatmaps from LiDAR and radar show nearly no indication, confirming the property-aware ability.}
    \label{fig:property_heatmap}
\end{figure}

\textbf{Property-aware response.}
Fig.~\ref{fig:property_heatmap} contrasts the BEV feature responses elicited by each sensor. The three modalities exhibit markedly different spatial statistics: camera activations are dense but confined to the frontal frustum, LiDAR yields broad and geometrically faithful coverage, and radar is extremely sparse with an order-of-magnitude smaller response magnitude (note the reduced color scale). Despite this imbalance, the fused response concentrates on the referred object, indicating that the soft-gated arbitration preserves the sparse yet critical radar evidence instead of letting the dense camera stream dominate, precisely the failure mode exhibited by naive fusion in Table~\ref{tab:experiment_benchmark}.

\begin{figure}
    \centering
    \includegraphics[width=0.99\linewidth]{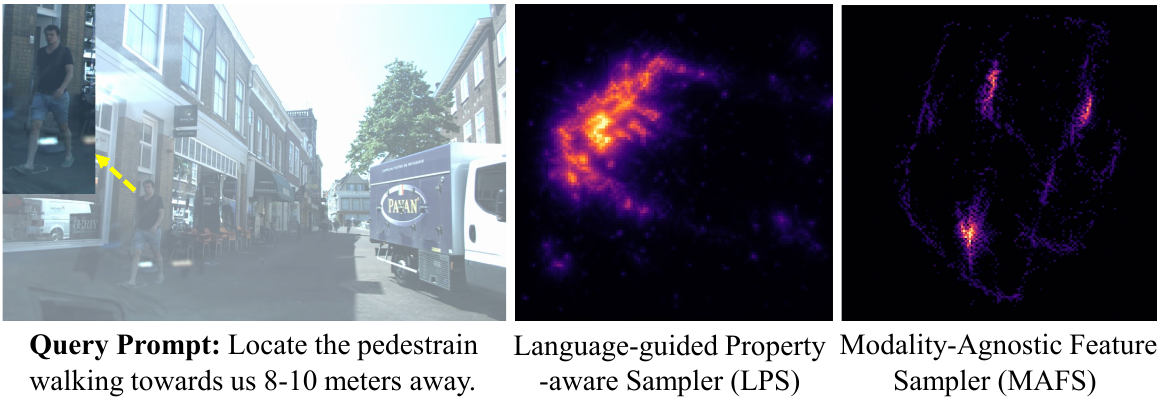}
    \vspace{-4mm}
    \caption{Sampling response of the proposed Language-Routed Property Sampler (LRPS) versus the Modality-Agnostic Feature Sampler (MAFS). Under the same prompt, LRPS concentrates its sampling on the referred pedestrian, whereas MAFS disperses across background and non-referred objects.}
    \label{fig:sample_heatmap}
\end{figure}

\textbf{Language-guided sampling.}
Fig.~\ref{fig:sample_heatmap} compares the sampling behavior of LRPS against the MAFS under an identical prompt. Guided by the query-level text modulation, LRPS concentrates its sampling density on the referred pedestrian, whereas MAFS disperses across background structures and non-referred objects. This gives direct visual evidence for the coarse text-conditioned anchoring of LRPS and explains its consistent margin over MAFS in Table~\ref{tab:sampler_comparison}.

\begin{figure}
    \centering
    \includegraphics[width=0.99\linewidth]{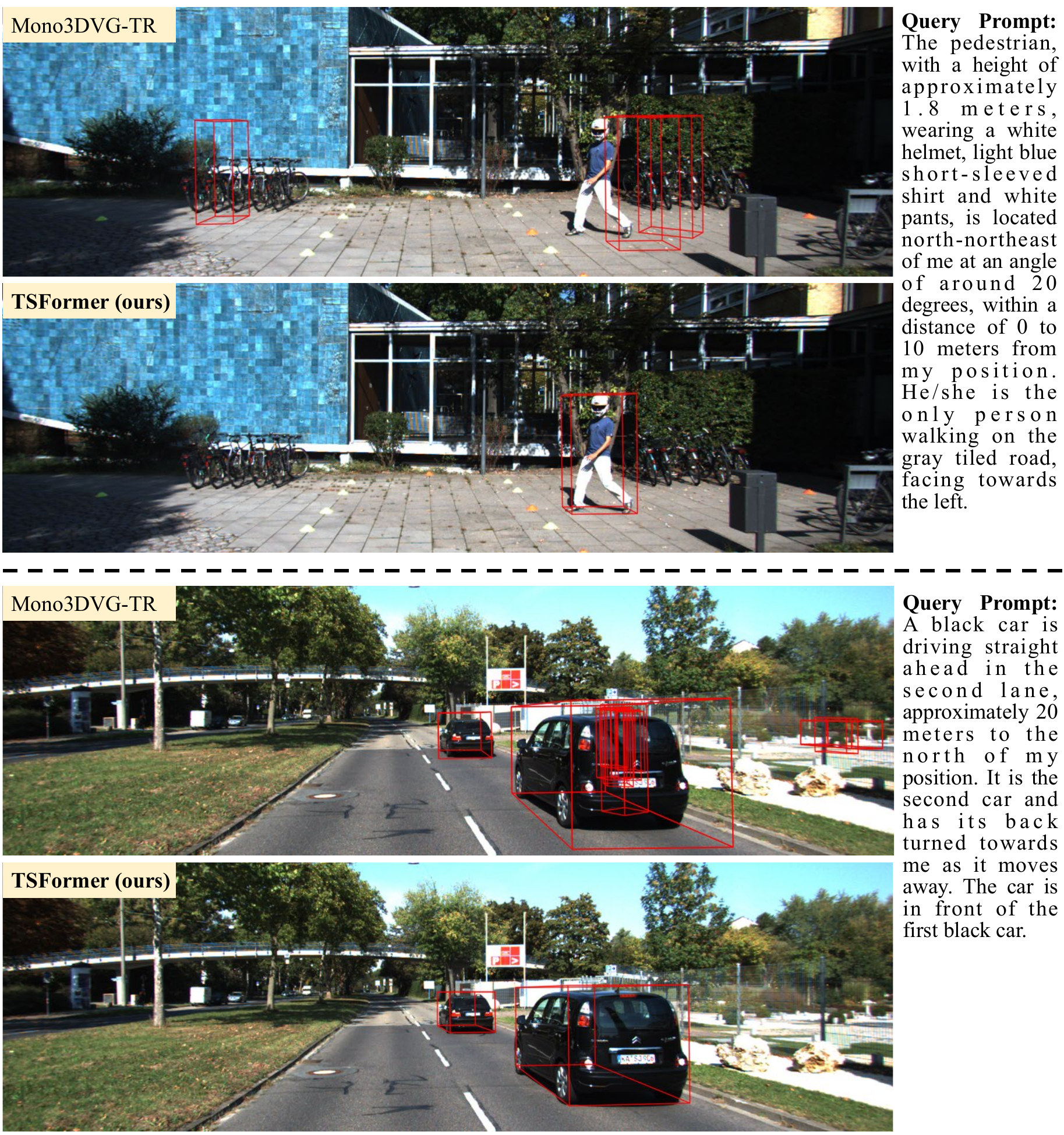}
    \vspace{-3mm}
    \caption{Qualitative comparison with the specialized monocular Mono3DVG-TR on the Mono3DRefer benchmark. Under identical query prompts, Mono3DVG-TR yields redundant and loosely aligned boxes, whereas TSFormer localizes the referred instance with a single, tightly fitted 3D box.}
    \label{fig:mono3dvg_vis}
\end{figure}

\textbf{Cross-dataset qualitative comparison.}
Fig.~\ref{fig:mono3dvg_vis} compares TSFormer against the specialized monocular Mono3DVG-TR on Mono3DRefer. For the pedestrian query, Mono3DVG-TR scatters several redundant, loosely localized boxes across the bicycle rack and the wrong region, whereas TSFormer isolates a single tightly fitted box on the referred pedestrian walking on the tiled road. The same pattern recurs on the ``second black car'' query, where Mono3DVG-TR produces overlapping, coarsely aligned boxes while TSFormer recovers a clean, well-oriented 3D box on the correct instance. These cases visually corroborate the generalization result in Table~\ref{tab:mono3drefer}: the property-aware design produces more precise 3D localization, which is exactly what the strict $\text{IoU}\!=\!0.5$ metric rewards.

\section{Conclusion}
\label{sec:conclusion}
In this work, we addressed the largely overlooked role of sensor-specific physical properties in 3D visual grounding. We introduced \textbf{Talk2Sensors}, the first tri-sensor (camera, LiDAR, and 4D mmWave radar) grounding benchmark, whose 8,682 prompts and 20,558 objects explicitly align natural language with heterogeneous appearance, geometry, and kinematic cues. Built upon it, we proposed \textbf{TSFormer}, a unified language-guided framework that couples a Language-Routed Property Sampler for coarse text-conditioned feature retrieval with a Sparse-Preserving Modality Arbiter module for fine-grained modality arbitration, routing each query to its most informative sensors while preventing dense modalities from overwhelming sparse but critical signals. Extensive experiments demonstrate that TSFormer establishes state-of-the-art performance under every sensor configuration, degrades gracefully under modality dropout, and generalizes to the monocular Mono3DRefer benchmark. In future work, we plan to extend the framework to adverse-weather and nighttime scenarios, where the complementary robustness of radar is expected to be even more pronounced, and to incorporate temporal reasoning for continuous kinematic grounding.


%

\appendices




\ifCLASSOPTIONcaptionsoff
  \newpage
\fi



\normalem
\footnotesize
\bibliographystyle{IEEEtran}
\bibliography{refs}
%

%








\end{document}